\DeclareRobustCommand\onedot{\futurelet\@let@token\@onedot}
\def\@onedot{\ifx\@let@token.\else.\null\fi\xspace}

\def\eg{\emph{e.g.}}

\def\etal{\emph{et al.}}

\documentclass[10pt,journal,cspaper,compsoc]{IEEEtran}

\ifCLASSOPTIONcompsoc
   \usepackage[nocompress]{cite}
\else
\fi
\usepackage{graphicx}
\usepackage{times}
\usepackage{epsfig}
\usepackage{graphicx}
\usepackage{amsmath}
\usepackage{amssymb}
\usepackage{color}
\usepackage{boxedminipage}
\usepackage{pifont}
\usepackage{subfigure}
\usepackage{flushend}
\usepackage{colortbl}
\usepackage{multirow}

\begin{document}
\title{Human-Machine CRFs for Identifying Bottlenecks in Holistic Scene Understanding}

\author{Roozbeh~Mottaghi, Sanja~Fidler, Alan~Yuille, Raquel~Urtasun, Devi~Parikh
\IEEEcompsocitemizethanks{\IEEEcompsocthanksitem R. Mottaghi is affiliated with Stanford University. A. Yuille is affiliated with UCLA. S. Fidler and R. Urtasun are with TTIC, and D. Parikh is with the Department of Electrical and Computer Engineering at Virginia Tech.}
}

\IEEEcompsoctitleabstractindextext{%

\begin{abstract}
Recent trends in image understanding have pushed for holistic scene understanding models that jointly reason about various tasks such as object detection, scene recognition, shape analysis, contextual reasoning, and local appearance based classifiers. In this work, we are interested in understanding the roles of these different tasks in improved scene understanding, in particular semantic segmentation, object detection and scene recognition. Towards this goal, we ``plug-in'' human subjects for each of the various components in a state-of-the-art conditional random field model. Comparisons among various hybrid human-machine CRFs give us indications of how much ``head room'' there is to improve scene understanding by focusing research efforts on various individual tasks. 
\end{abstract}

}


\newcommand{\roozbeh}{\textcolor{blue}}
\newcommand{\sanja}{\textcolor{green}}
\newcommand{\raquel}{\textcolor{magenta}}
\newcommand{\devi}{\textcolor{red}}
\newcommand{\change}{\textcolor{red}}

\definecolor{orange}{rgb}{1,0.9,0.65}
\definecolor{gr}{rgb}{0.9,1,0.6}
\definecolor{bl}{rgb}{0.9,0.8,1}
\definecolor{bg}{rgb}{0.8,0.9,1}
\definecolor{orr}{rgb}{1,0.85,0.4}
\definecolor{grr}{rgb}{0.8,1,0.5}
\definecolor{blr}{rgb}{0.85,0.45,1}
\definecolor{bgr}{rgb}{0.5,0.9,1}
\definecolor{gry}{rgb}{0.92,0.92,0.92}
\definecolor{rr}{rgb}{1,0.85,0.9}
\definecolor{blb}{rgb}{0.78,0.84,1}

\maketitle

\IEEEdisplaynotcompsoctitleabstractindextext

%
\IEEEpeerreviewmaketitle

\vspace{-10pt}
\section{Introduction}
Automatic holistic scene understanding is one of the holy grails of computer vision. Given the lofty challenge it presents, the community has historically studied individual tasks in isolation. This includes tasks such as object detection~\cite{s:felzenswalb10}, scene recognition~\cite{s:xiao10}, contextual reasoning among objects~\cite{Rabinovich}, and pose estimation \cite{Deva}. However, clearly these tasks are related. For example, knowing that the image is a street scene influences where and at what scales we expect to find people. Detecting a microwave in an image can help identify a kitchen scene. Studies have shown that humans can effectively leverage contextual information from the entire scene to recognize objects in low resolution images that can not be recognized in isolation~\cite{ref:tiny_images}. In fact, different and functionally complementary regions in the brain are known to co-operate to perform scene understanding~\cite{Aude}. 

Recent works~\cite{s:gould09,Jian, s:heitz08, li10}, have thus pushed on \emph{holistic} scene understanding models. The advent of general learning and inference techniques for graphical models has provided the community with appropriate tools to allow for joint modeling of various scene understanding tasks. These have led to some of the state-of-the-art approaches.  

In this paper, we aim to determine the relative importance of the different recognition tasks in aiding holistic scene understanding. We wish to know, which of the tasks if improved, can boost performance significantly. In other words, to what degree can we expect to improve holistic understanding performance by improving the performance of individual tasks? We argue that understanding which problems to solve is as important as determining how to solve them. Such an understanding can provide valuable insights into which research directions to pursue for further improving the state-of-the-art. We use semantic segmentation, object detection and scene recognition accuracies as proxies for holistic scene understanding performance.

We analyze one of the most recent and comprehensive holistic scene understanding models \cite{Jian}. It is a conditional random field (CRF) that models the interplay between a variety of factors such as local super-pixel appearance, object detection, scene recognition, shape analysis, class co-occurrence, and compatibility of classes with scene categories. To gain insights into the relative importance of these different factors or tasks, we isolate each task, and substitute a machine with a human for that task, keeping the rest of the model intact. The resultant improvement in  performance of the model, if any, gives us an indication of how much ``head room'' there is to improve performance by focusing research efforts on that task. Note that human outputs are \emph{not} synonymous with ground truth information, because the tasks are performed in isolation. For instance, humans would not produce ground truth labels when asked to classify a super-pixel in isolation into one of several categories\footnote{Of course, ground truth segmentation annotations are themselves generated by humans, but by viewing the whole image and leveraging information from the entire scene. In this study, we are interested in evaluating how each recognition task in \emph{isolation} can help the overall performance.}. In fact, because of inherent local ambiguities, the most intelligent machine of the future will likely be unable to do so either. Hence, the use of human subjects in our studies is key, as it gives us a \emph{feasible} point (hence, a lower- bound) of what can be done.


Our slew of studies reveal several interesting findings. For instance, we found that human classification of \emph{isolated} super-pixels when fed into the model provides a 5\% improvement in segmentation accuracy on the MSRC dataset. Hence, research efforts focussed towards the specific task of classifying super-pixels in isolation may prove to be fruitful. Even more intriguing is that the human classification of super-pixels is in fact less accurate than machine classification. However when plugged into the holistic model, human potentials provide a significant boost in performance. This indicates that to improve segmentation performance, instead of attempting to build super-pixel classifiers that make fewer mistakes, research efforts should be dedicated towards making the right kinds of mistakes. This provides a refreshing new take on the now well studied semantic segmentation task. 

Excited by this insight, we conducted a thorough analysis of the human generated super-pixel potentials to identify precisely how they differ from existing machine potentials. Our analysis inspired a rather simple modification of the machine potentials which resulted in a significant increase of 2.4\% in the machine accuracy (i.e. no human involvement) over the state-of-the-art on the MSRC dataset. Additionally, we annotated a subset of PASCAL dataset with 14 background classes. Our human studies and machine experiments on PACSAL reveal similar complementary patterns in the mistakes made by humans and machines. 

In addition, through a series of human studies and machine experiments on MSRC we show that a reliable object shape model is beneficial for semantic segmentation and object detection. We demonstrate that humans are not good at deciphering object shape from state-of-art UCM segment boundaries any more than existing machine approaches.

We also studied how well humans can leverage the contextual information modeled in the CRF. We measure human and machine segmentation performance while progressively increasing the amount of contextual information available. We find that even though humans perform significantly worse than machines when classifying isolated super-pixels, they perform better than machines when both are given access to the contextual information modeled by the CRF.

Section~\ref{sec:related} describes other holistic scene understanding works and also human studies related to computer vision models. In Section~\ref{sec:mac_approach}, we explain the machine CRF model used for our experiments. In Section~\ref{sec:dataset}, we explain the choice of the datasets. Section~\ref{sec:pots} explains how we obtain the machine and human-based potential functions for the CRF model. Section~\ref{sec:results} presents the result of plugging-in human or ground truth potential functions in the CRF model. Finally, in Section~\ref{sec:modeljust} we analyze the model to see how much potential it holds for improvement and whether the components used in the model are beneficial for humans.


\section{Related Work}
\label{sec:related}
\noindent \textbf{Holistic Scene Understanding:} The key motivation behind holistic scene understanding, going back to the seminal work of Barrow in the seventies~\cite{s:barrow78}, is  that ambiguities in visual information can only be resolved when many visual processes are working collaboratively. A variety of holistic approaches  have since been proposed. Many of these works incorporate various tasks in a sequential fashion, by using the output of one task (e.g., object detection) as features for other tasks (e.g., depth estimation, object segmentation)~\cite{s:hoiem08,s:heitz08,Lempitsky09,Brox11,s:gu09}.  There are fewer efforts on joint reasoning of the various recognition tasks. In \cite{s:torralba05}, contextual information was incorporated into a CRF leading to improved object detection. A hierarchical generative model spanning parts, objects and scenes is learnt in~\cite{s:sudderth05}. Joint estimation of depth, scene type, and object locations is performed in~\cite{li10}. Spatial contextual interactions between objects have also been modeled~\cite{s:kumar05,Rabinovich}. Image segmentation and object detection are jointly modeled in~\cite{s:ladicky10,s:wojek08,s:gould09} using a CRF. \cite{s:gonfaus10} also models global image classification in the CRF. In this paper, orthogonal to these advances, we propose the use of human subjects to understand the relative importance of various recognition tasks in aiding holistic scene understanding. 

\noindent \textbf{Human-Studies:}
Numerous human-studies have been conducted to  understand the human ability to segment an image into meaningful regions or objects. Rivest and Cavanagh \cite{Rivest1996} found that luminance, color, motion and texture cues for contour detections are integrated at a common site in the brain. Fowlkes \cite{FowlkesThesis} found that machine performance at detecting boundaries is equivalent to human performance in small gray-scale patches. These and other studies are focused on the problem of unsupervised segmentation, where the task is to identify object boundaries.  In contrast, we are interested in holistic scene understanding, including the  task of identifying the semantic category of each pixel in the image.


Several works have studied high-level recognition tasks in humans. Fei-Fei~\etal~\cite{ref:rapid_no_attention} show that humans can recognize scenes rapidly even while being distracted.  Bachmann~\etal~\cite{ref:tiny_faces} show that humans can reliably recognize faces in $16 \times 16$ images, and Oliva~\etal~\cite{ref:low_res_scene} present similar results for scene recognition. Torralba~\etal~\cite{ref:tiny_images} show that humans can reliably detect objects in $32 \times 32$ images. In contrast, we study human performance at tasks that closely mimic existing holistic computational models for holistic scene understanding in order to identify bottlenecks, and better guide future research efforts. 

\emph{Human debugging} i.e. using human subjects to identify bottlenecks in existing computer vision systems has been recently explored for a number of different applications such as analyzing the relative importance of features, amount of training data and choice of classifiers in image classification~\cite{Parikh10}, of part detection, spatial modeling and non-maximal suppression in person detection~\cite{Parikh2011}, of local and global image representations in image classification~\cite{Parikh_jum}, and of low-, mid- and high-level cues in detecting object contours~\cite{Zitnick12}. In this work, we are interested in systematically analyzing the roles played by several high- and mid-level tasks such as grouping, shape analysis, scene recognition, object detection and contextual interactions in \emph{holistic scene understanding}. While similar at the level of exploiting human involvement, the problem, the model, the methodologies of the human studies and machine experiments, as well as the findings and insights are all novel.


\section{CRF Model} 
\label{sec:mac_approach}

We analyze the recently introduced CRF model of~\cite{Jian} which achieved state-of-the-art performance on the MSRC dataset by reasoning jointly about a variety of scene components. While the model shares similarities with past work~\cite{s:ladicky09,s:ladicky10,s:gonfaus10}, we choose this model because it  provides state-of-the-art performance in holistic scene understanding, and thus forms a great starting point to ask ``which components need to be improved to push the state-of-the-art further?''. Moreover, it has a simple ``plug-and-play'' architecture making it feasible to insert humans in the model. Inference is performed via message passing \cite{s:schwing11} and so it places no restrictions (e.g. submodularity) on the potentials. This allows us to conveniently replace the machine potentials with human responses: after all, we cannot quite require humans to be submodular! 

\begin{figure}
\centering
\includegraphics[width=0.9\columnwidth]{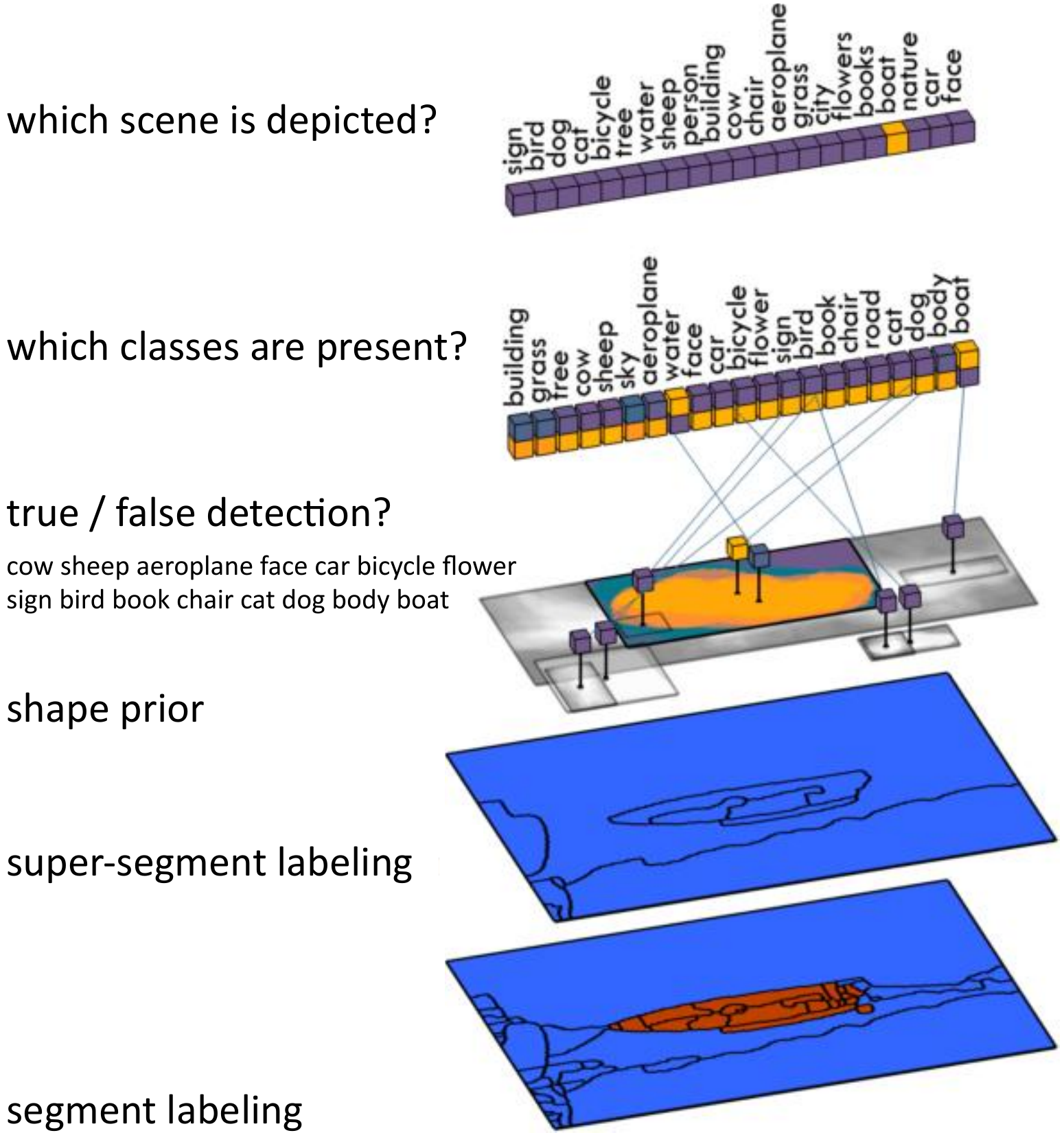}
\caption{Overview of the holistic scene model of~\cite{Jian} that we analyze using human subjects. For clarity, not all connections in the model are shown here.}
\label{fig:model}
\end{figure}

We now briefly review this model (Figure~\ref{fig:model}). We refer the reader to \cite{Jian} for further technical details. The problem of holistic scene understanding is formulated as that of inference in a CRF. The random field contains variables representing the class labels of image segments at two levels in a segmentation hierarchy: super-pixels and larger segments. To be consistent with~\cite{Jian}, we will refer to them as segments and super-segments. The model also has binary variables indicating the correctness of candidate object detection bounding boxes. In addition, a multi-labeled variable represents the scene type and binary variables encode the presence/absence of a class in the scene. 


The segments and super-segments reason about the semantic class labels to be assigned to each pixel in the image. The model employs these two segmentation layers for computational efficiency: the super-segments are fewer but more densely connected to other parts of the model. The binary variables corresponding to each candidate bounding box generated by an object detector allow the model to accept or reject these detections. A shape prior is associated with these nodes encouraging segments to take on corresponding class labels. The binary class variables reason about which semantic classes are present in the image. This allows for a natural way to model class co-occurrences as well as scene-class affinities. These binary class variables are connected to i) the super-segments via a consistency potential that ensures that the binary variables are turned on if a super-segment takes the corresponding class label ii) binary detector variables via a similar consistency potential iii) the scene variable via a potential that encourages certain classes to be present in certain scene types iv) to each other via a potential that encourages certain classes to co-occur more than others. 

More formally, let $x_i \in \{1, \cdots, C \}$ and  $y_j \in \{1, \cdots, C \}$  be two random variables representing the class label of the $i$-th segment and $j$-th super-segment. We represent candidate detections as binary random variables,   $b_i \in \{0,1\}$,   taking value  $0$ when the detection is a false detection. A part-based mixture model \cite{s:felzenswalb10} is used to generate candidates. The detector provides us with an object class ($c_i$), the score ($r_i$), the location and aspect ratio of the bounding box, as well as the root mixture component ID that has generated the detection  ($m_i$). The latter gives us information about the expected shape of the object. Let $z_k \in \{0,1\}$ be a random variable which takes value $1$ if class $k$ is present in the image, and let  $s \in \{1,\dots, C_l\}$ be a random variable representing the scene type among $C_l$ possible candidates. The parameters corresponding to different potential terms in the model are learnt in a discriminative fashion \cite{s:hazan10}. Before we provide details about how the various machine potentials are computed, we first discuss the dataset we work with to ground further descriptions. 

\section{Dataset} 
\label{sec:dataset}

We use the standard MSRC-21 \cite{s:shotton09} semantic labeling benchmark, also used by~\cite{Jian},  as it contains ``stuff'' (e.g.,  \emph{sky, water}) as well as ``things'' (i.e., shape-defined classes such as \emph{cow, car}). 
The PASCAL dataset is more challenging in terms of object (``things'') detection and segmentation, but a large portion of its images, especially ``stuff'', is unlabeled. Hence, we augment a subset of PASCAL dataset with additional categories and report results on that. 

We use the more precise ground truth of MSRC provided by Malisiewicz and Efros~\cite{j:Malisiewicz} and used in~\cite{Jian}, as it offers  a more accurate measure of performance. We use the same scene category and object detection annotations as in~\cite{Jian}. Table~\ref{tab:msrcinfo} lists this information. As the performance metric we use average per-class recall (average accuracy). Similar trends in our results hold for average per-pixel recall (global accuracy~\cite{s:ladicky10}) as well.  We use the standard train/test split from \cite{s:shotton08} to train all machine potentials, described next.




\begin{table*}[t]
{\scriptsize
\begin{center}
\addtolength{\tabcolsep}{-5pt}
\begin{tabular}{|ccccccccccccccccccccc|ccccccccccccccccccccc|ccccccccccccccc|}
\multicolumn{21}{>{\columncolor{bl}}c|}{{\bf semantic labels}} & \multicolumn{21}{>{\columncolor{orange}}c|}{{\bf scene types}} & \multicolumn{15}{>{\columncolor{gr}}c|}{{\bf obj. detection classes}}\\
\hline
$\,$\rotatebox{90}{building} & \rotatebox{90}{grass} & \rotatebox{90}{tree} & \rotatebox{90}{cow} & \rotatebox{90}{sheep} & \rotatebox{90}{sky} & \rotatebox{90}{aeroplane$\ $} & \rotatebox{90}{water} & \rotatebox{90}{face} & \rotatebox{90}{car} & \rotatebox{90}{bicycle} & \rotatebox{90}{flower} & \rotatebox{90}{sign} & \rotatebox{90}{bird} & \rotatebox{90}{book} & \rotatebox{90}{chair} & \rotatebox{90}{road} & \rotatebox{90}{cat} & \rotatebox{90}{dog} & \rotatebox{90}{body} & \rotatebox{90}{boat}$\;$ & 
 \rotatebox{90}{sign} & \rotatebox{90}{bird} & \rotatebox{90}{dog} & \rotatebox{90}{cat} & \rotatebox{90}{bicycle} & \rotatebox{90}{tree} & \rotatebox{90}{water} & \rotatebox{90}{sheep} & \rotatebox{90}{person} & \rotatebox{90}{building} & \rotatebox{90}{cow} & \rotatebox{90}{chair} & \rotatebox{90}{aeroplane} & \rotatebox{90}{grass} & \rotatebox{90}{city} & \rotatebox{90}{flowers} & \rotatebox{90}{books} & \rotatebox{90}{boat} & \rotatebox{90}{nature} & \rotatebox{90}{car} & \rotatebox{90}{face}$\;$ &
$\,$ \rotatebox{90}{cow} & \rotatebox{90}{sheep} & \rotatebox{90}{aeroplane} & \rotatebox{90}{face} & \rotatebox{90}{car} & \rotatebox{90}{bicycle} & \rotatebox{90}{flower} & \rotatebox{90}{sign} & \rotatebox{90}{bird} & \rotatebox{90}{book} & \rotatebox{90}{chair} & \rotatebox{90}{cat} & \rotatebox{90}{dog} & \rotatebox{90}{body} & \rotatebox{90}{boat}$\;$ \\
\hline
\end{tabular}
\end{center}
}
\caption{MSRC-21 dataset information}
\label{tab:msrcinfo}
\end{table*}

\section{Machine \& Human CRF Potentials}
\label{sec:pots}
We now describe the machine and human potentials we employed. Section~\ref{sec:results} presents the results of feeding the human ``potentials'' into the machine model. Our choices for the machine potentials closely follow those made in~\cite{Jian}. For human potentials, we performed all human studies on Amazon Mechanical Turk. Unless specified otherwise, each task was performed by 10 different subjects. Depending on the task, we paid participants $3-5$ cents for answering $20$ questions. The response time was fast, taking $1$ to $2$ days to perform each experiment. We randomly checked the responses of the workers and excluded those that did not follow the instructions\footnote{As our experiments will demonstrate, the fact that we can train the CRF parameters on responses of human subjects and have it generalize well to human responses to held out test images vouches for the reliability of the collected human responses.}. More than 500 subjects participated in our studies that involved $\sim 300,000$ crowd-sourced tasks, making the results obtained likely to be fairly stable across a different sampling of subjects. 

\subsection{Segments and super-segments} 

\noindent \textbf{Machine:} We utilize UCM~\cite{Arbelaez11} to create our segments and super-segments as   it returns a small number of segments that tend to respect the true object boundaries well. We use thresholds $0.08$ and $0.16$ for the segments and super-segments respectively. On average, this results in $65$ segments and $19$ super-segments per image for the MSRC dataset. We use the output of the modified Textoonboost~\cite{s:shotton09} in \cite{s:ladicky09} to get pixel-wise potentials and average those within the segments and super-segments to get the unary potentials. Following~\cite{Kohli07}, we connect the two levels via a pairwise $P^n$ potential that encourages segments and super-segments to take the same label. 

\begin{figure}[tp]
\centering
\includegraphics[width=1\linewidth]{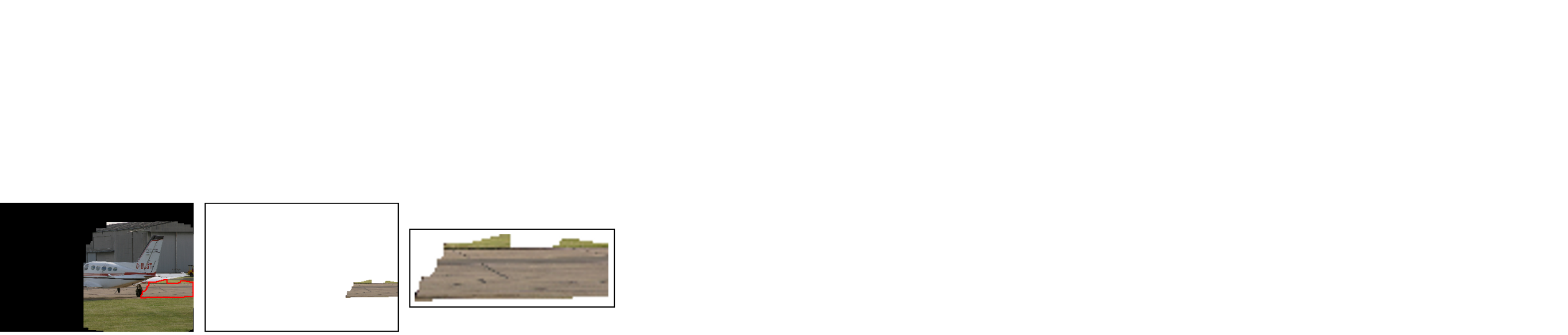}
\caption{Human segment labeling interface. Left panel: Human subjects are asked to classify the segment outlined with a red boundary, where they can see all the information that the machine classifier has access to. Middle panel: Only the segment is shown to the human subjects. Right panel: Location and scale information is discarded when the segment is shown to the subjects.}
\label{fig:sp_interfaces}
\end{figure}

\noindent \textbf{Human:} 
The study involves having human subjects classify segments into one of the semantic categories. We experiment with three different visualizations, which are depicted in Figure~\ref{fig:sp_interfaces}. The first interface (left) shows all the information that the machine exploited when classifying a segment i.e. the image area in which the image features used by the classifier were extracted. Note however, that the machine employed a bag-of-words model to compute the machine potentials and does not exploit the relative location information. The machine classifier, TextonBoost~\cite{s:shotton09} in particular, has access to a large neighborhood (200x200 pixels) around the segment. Clearly, it does not use information only from the pixels in the segment while classifying the segment. However, showing all the information that the machine uses to human subjects would lead to nearly 100\% classification accuracy by the subjects, leaving us with little insights to gain. Therefore, we explored two other visualizations with impoverished information. The second interface (middle) only shows the pixels that belong to the segment. The third visualization (right) does the same, but discards the scale and location information of the segment, which  is shown at the center of the image and its largest dimension is resized to 240 pixels. We asked the subjects to classify 25 segments selected for each class at random  from the set of segments containing more than 500 pixels. 

The pixel-wise accuracies obtained via the three visualizations are 99.2\%, 79.0\%, and 75.6\%, respectively.  We chose the second interface for the rest of our experiments, which involved having subjects label all segments and super-segments from the MSRC dataset containing more than 500 pixels. This resulted in 10976 segments and 6770 super-segments. They cover 90.2\% and 97.7\% of all pixels in the dataset\footnote{Covering 100\% of the pixels in the dataset would involve labeling three times the number of segments, and the resources seemed better utilized in the other human studies.}.

\begin{figure}[tp]
\centering
\includegraphics[width=1.0\linewidth]{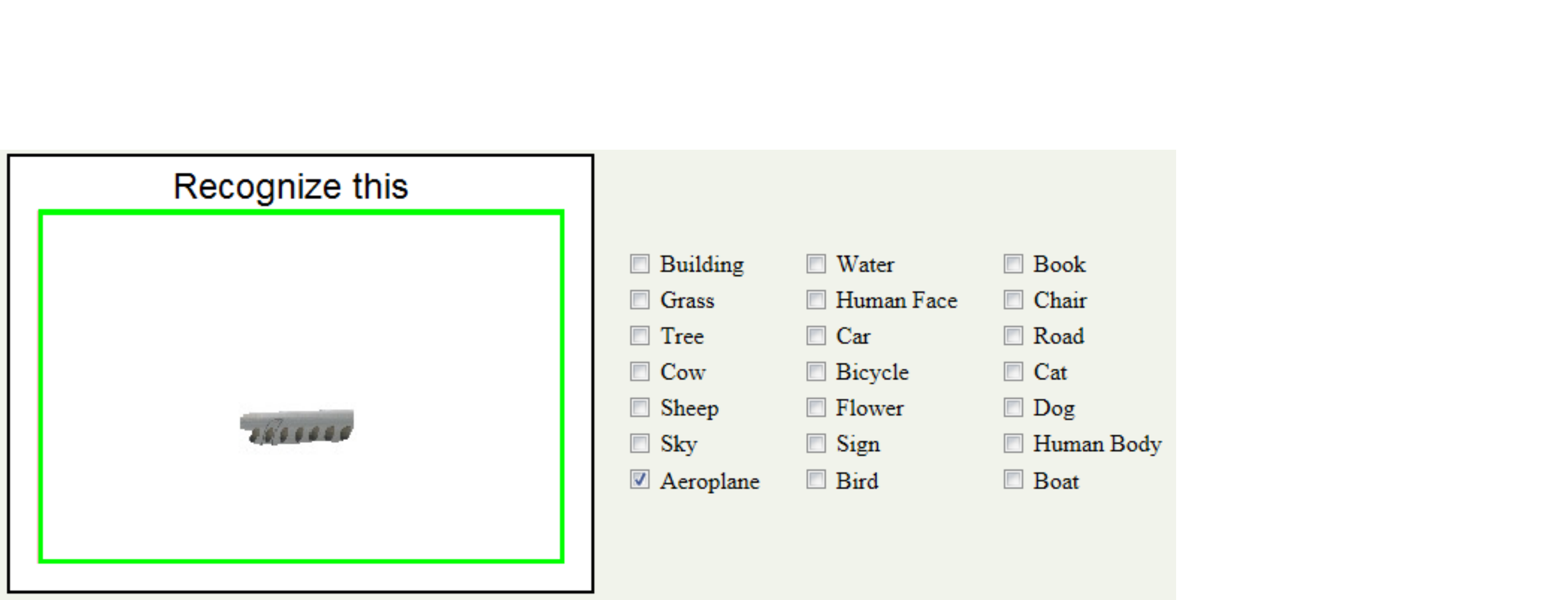}
\caption{Segment labeling interface. We ask the human subjects to choose the category that the segment belongs to. If the subjects are confused among a few categories, they have the option of choosing more than one answer.}
\label{fig:interf}
\end{figure}

We experimented with several interfaces e.g. showing subjects a collection of segments and asking them to click on all the ones likely to belong to a certain category, or allowing a subject to select only one category per segment, etc. before converging to the one that resulted in most consistent responses from subjects (Figure~\ref{fig:interf}) where subjects are asked to select all categories that a segment may belong to.

Note that a 200 x 200 window (used by the machine classifier) occupies nearly 60\% of the image. If this were shown to the human subjects, it would result in them potentially using holistic scene understanding while classifying the segments. This would contradict our goal of having humans perform individual tasks in isolation. More importantly, a direct comparison between humans and machines is not of interest to us. We are interested in understanding the potential each component in the model holds. To this goal, the discrepancy in information shown to humans and machines is not a concern, as long as humans are not shown \emph{more} information than the machine has access to. 

\def\IH{2.69cm}
\def\IHH{2.53cm}
\begin{figure*}[tp]
\centering
\begin{minipage}{0.9\linewidth}
\includegraphics[height=\IH]{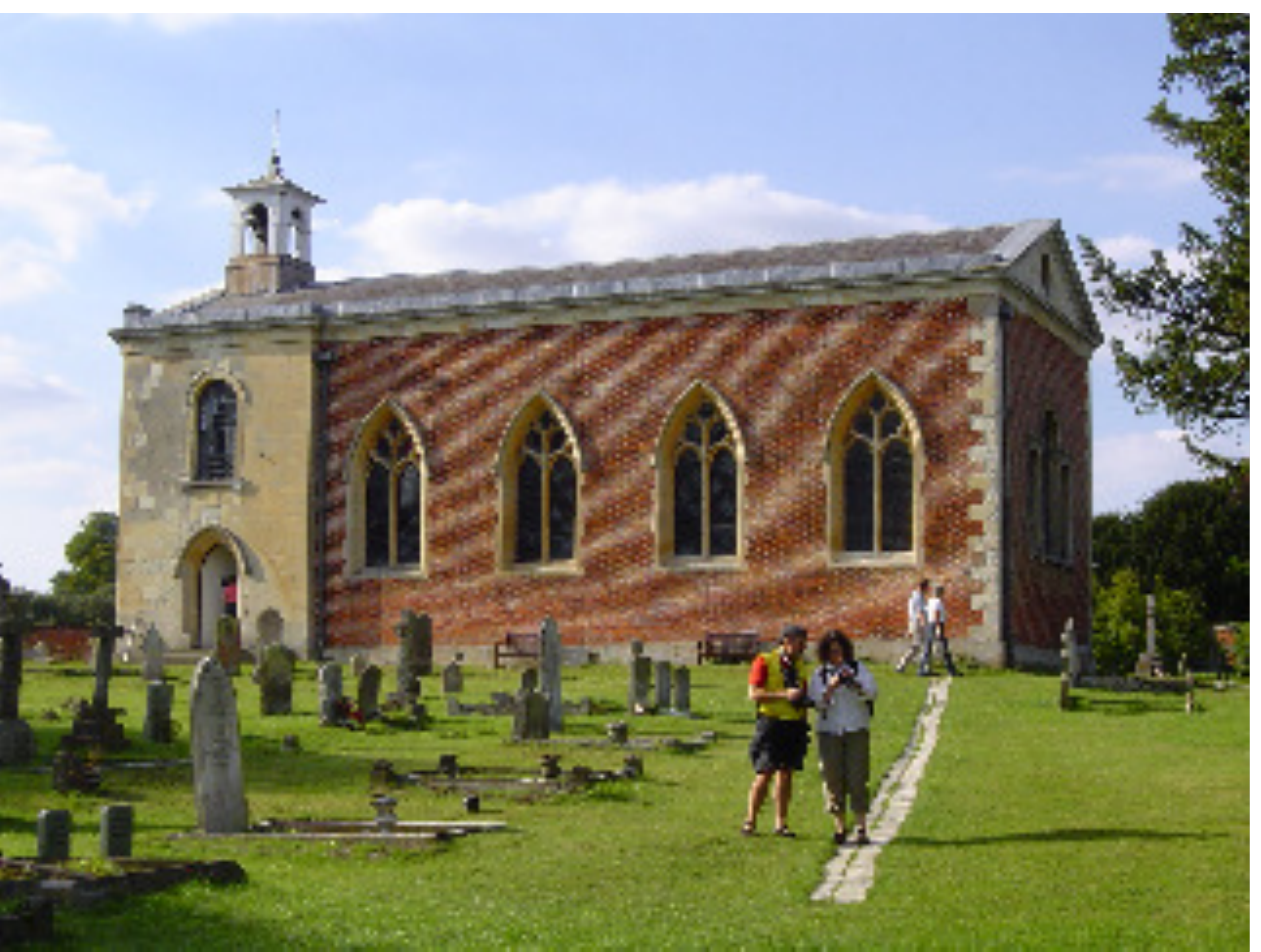}
\includegraphics[height=\IH]{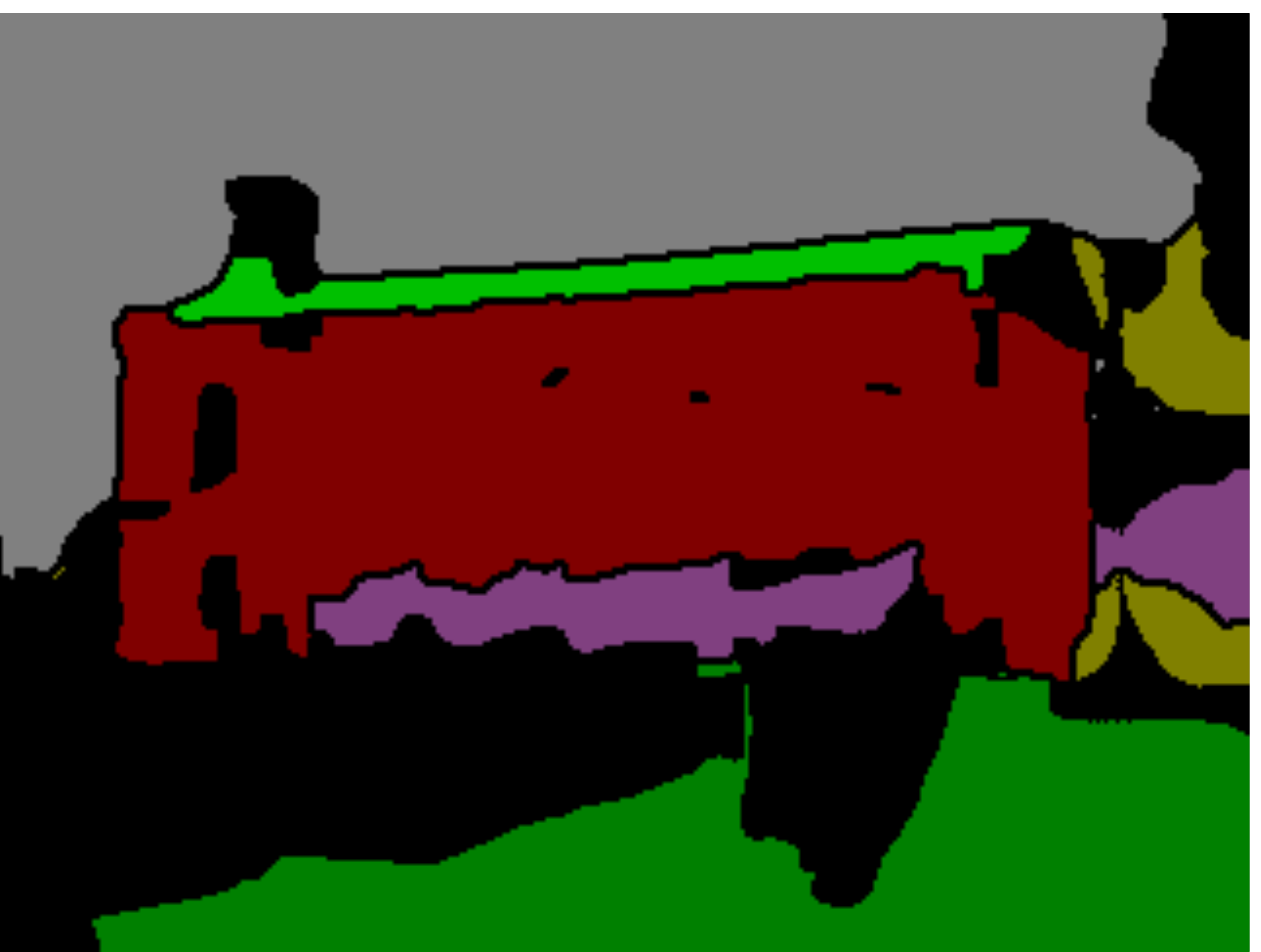}
\includegraphics[height=\IH]{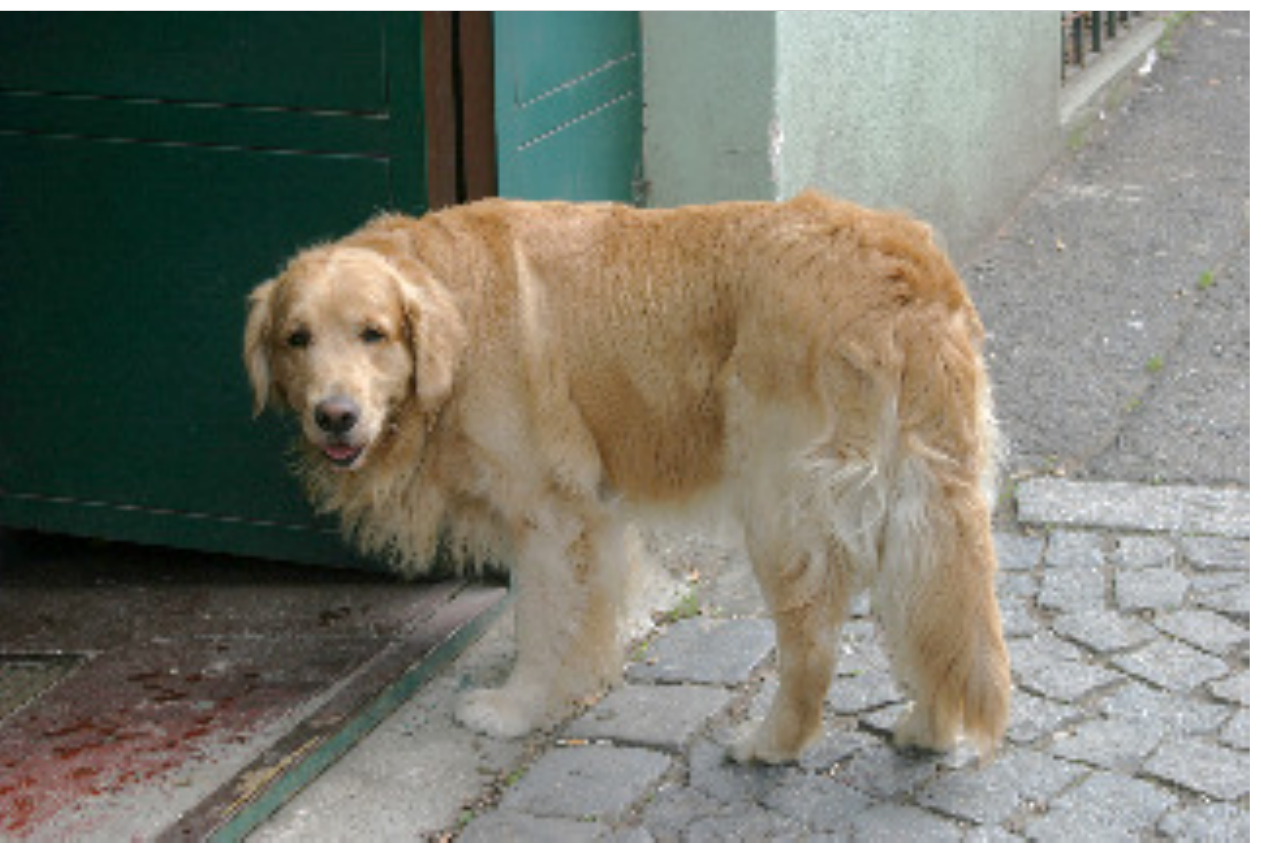}
\includegraphics[height=\IH]{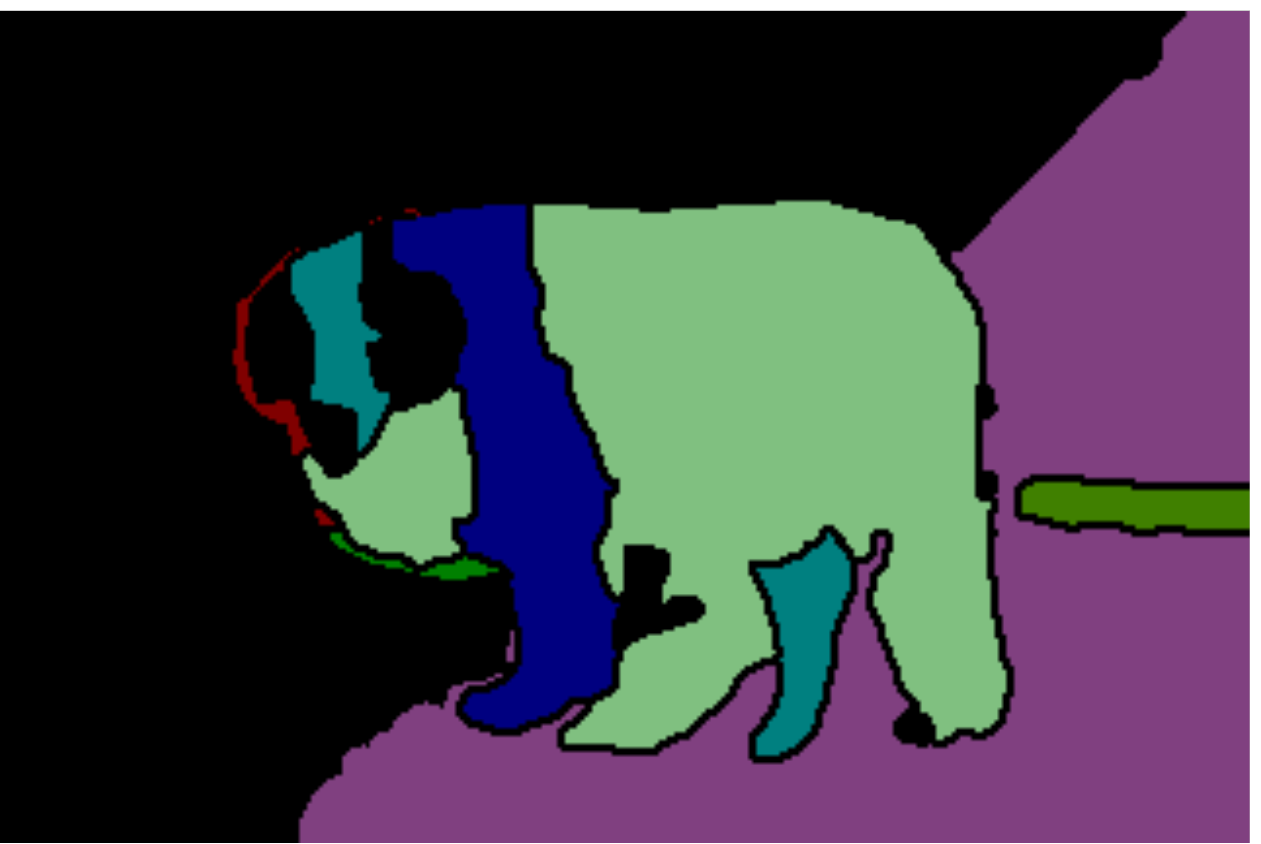}
\includegraphics[height=\IHH]{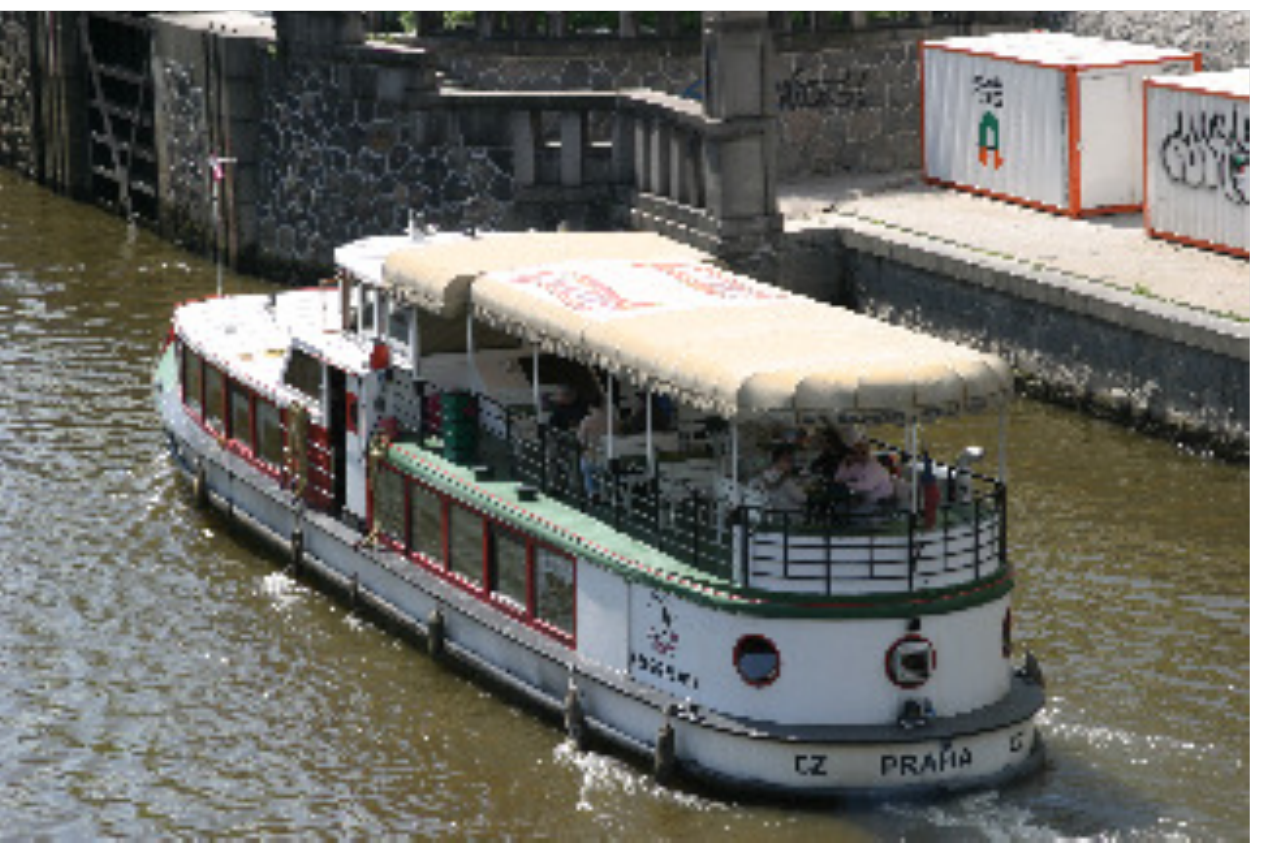}
\includegraphics[height=\IHH]{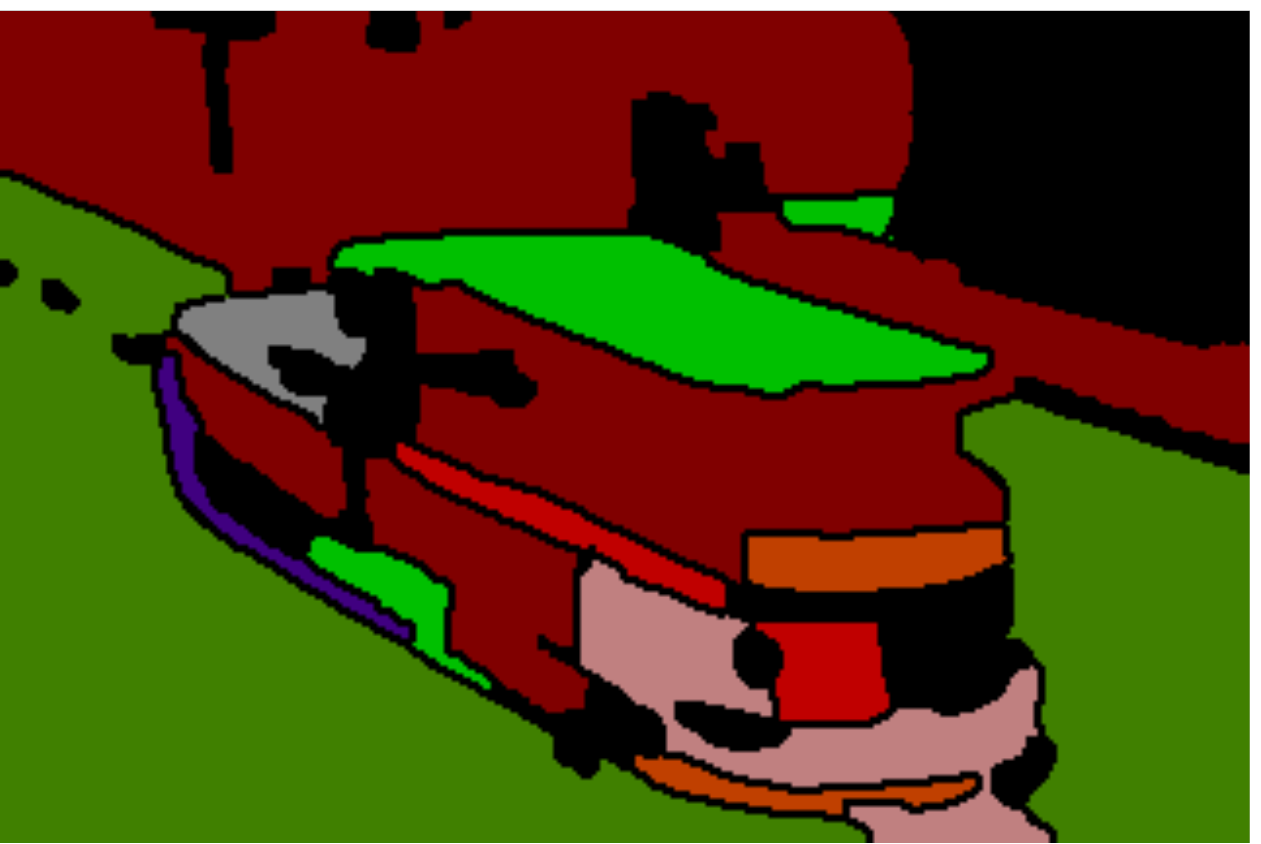}
\includegraphics[height=\IHH]{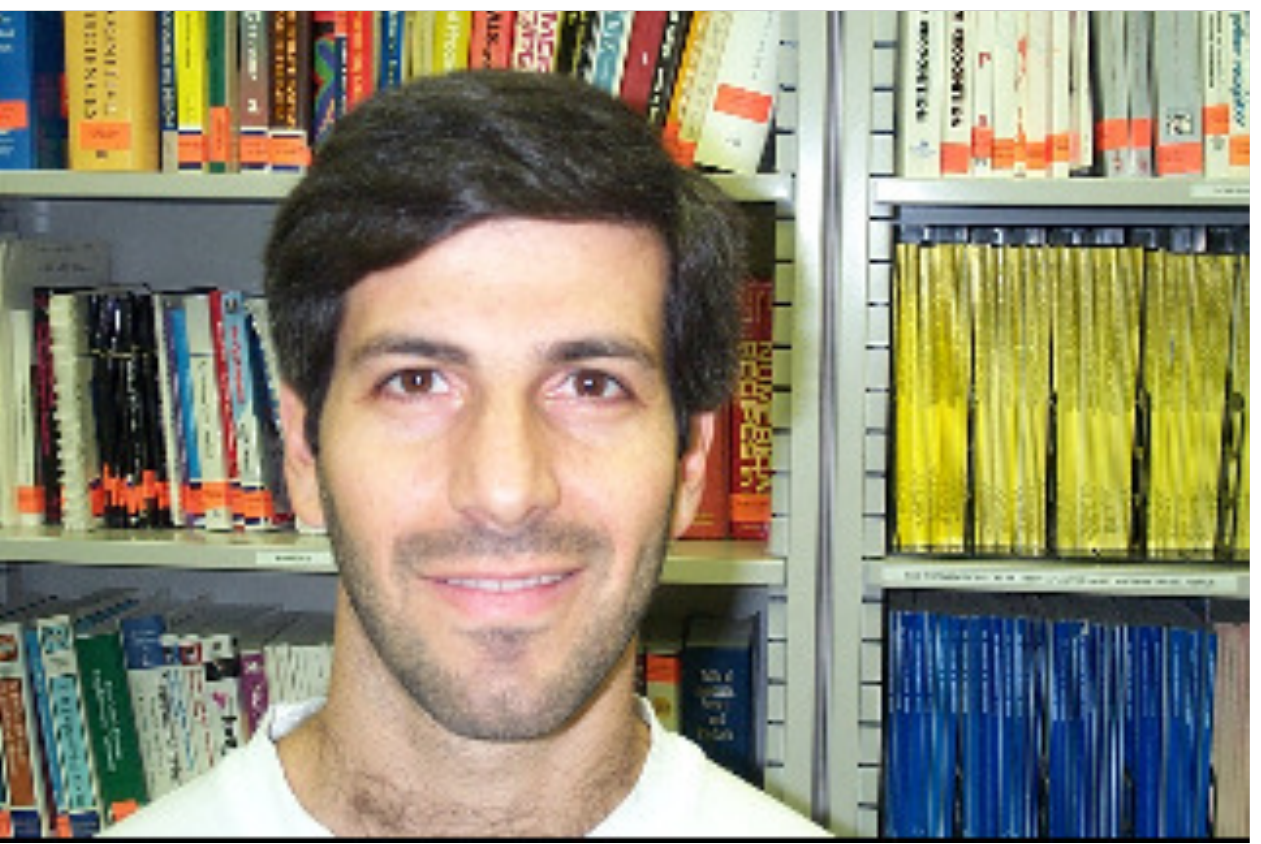}
\includegraphics[height=\IHH]{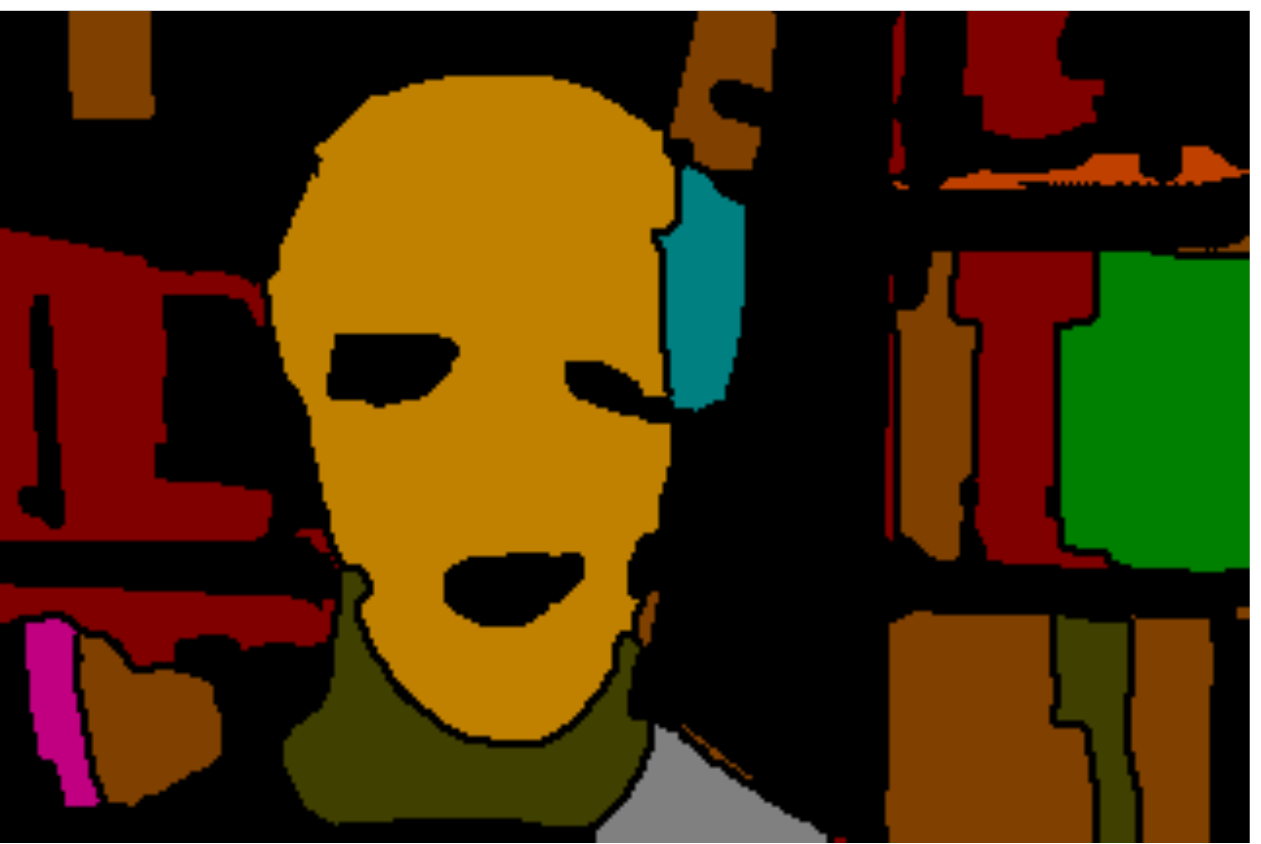}
\end{minipage}
\begin{minipage}{0.092\linewidth}
\includegraphics[width=1\linewidth,trim=0 0 10 0,clip=true]{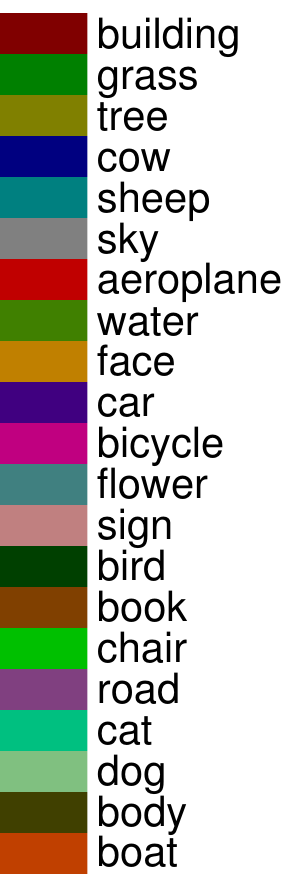}
\end{minipage}
\caption{Human labeling results on isolated segments.}
\label{fig:isolated}
\end{figure*}

Figure~\ref{fig:isolated} shows examples of segmentations obtained by assigning each segment to the class with most human votes. The black regions correspond to either the ``void'' class (unlabeled regions in the MSRC dataset) or to small segments not being shown to the subjects. Assigning each segment to the class with the highest number of human votes achieves an accuracy of 72.2\%, as compared to 77.4\% for machines\footnote{This accuracy is calculated only over segments larger than 500 pixels that were shown to humans. Machine accuracy over all segments is 74.2\%.}. Accuracy for super-segments is 84.3\% and 79.6\% respectively. As expected, humans perform rather poorly when only local information is available. However, they are better at classifying certain ``easy'', distinctive, classes or classes they are familiar with e.g., \emph{faces} (see confusion matrix in Figure~\ref{fig:cm_hum}). 

The $C$ dimensional human unary potential for a (super)segment is proportional to the number of times subjects selected each class, normalized to sum to 1. We set the potentials for the unlabeled (smaller than 500 pixels) (super)seg\-m\-ents to be uniform. 


\subsection{Class occurrence and co-occurrence} 

\noindent \textbf{Machine:} We use class-occurrence statistics extracted from training data as a unary potential on $z_k$. We also employ pairwise potentials between $z_i$ and $z_k$ that capture co-occurrence statistics of pairs of classes. However, for efficiency reasons, instead of utilizing a fully connected graph, we use a tree-structure obtained via the Chow-Liu algorithm~\cite{s:chow68} on the class-class co-occurrence matrix. 

\noindent \textbf{Human:} To obtain class-occurrence, we showed subjects 50 random images from the MSRC dataset to help them build an intuition for the image collection (not to count the occurrence of objects in the images). For all pairs of categories, we then ask subjects which category is more likely to occur in an image from the collection. We build the class unary potentials by counting how often each class was preferred over all other classes. We ask MAP-like questions (``which is more likely'') to build an estimate of the marginals (``how likely is this?'') because asking subjects to provide scalar values for the likelihood of something is prone to high variance and inconsistencies across subjects.

\begin{figure}[htp]
 \centering
 \subfigure[Human]{
 \includegraphics[scale=0.25,trim=20 86 10 80,clip=true]{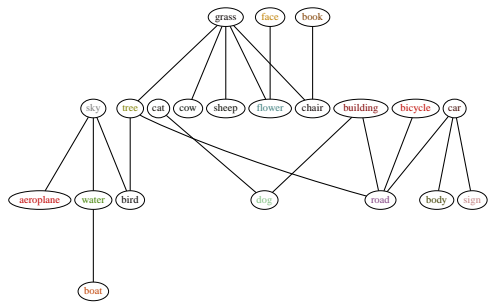}
 \label{fig:chowhum}
 }
 \subfigure[Machine]{
 \includegraphics[scale=0.25,trim=20 70 10 40,clip=true]{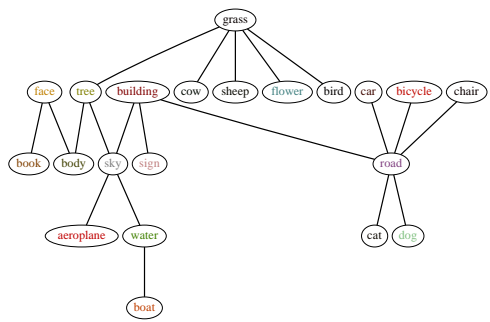}
 \label{fig:chomach}
 }

\caption{Chow-Liu trees for humans and machine. The trees share several similarities.}
\label{fig:chowliu}
\end{figure}

To obtain the human co-occurrence potentials we ask subjects the following question for all triplets of categories \{$z_i, z_j, z_k$\}: ``Which scenario is more likely to occur in an image? Observing ($z_i$ and $z_j$) or ($z_i$ and $z_k$)?''. Note that in this experiment we did not show subjects any images. The obtained statistics thus reflect human perception of class co-occurrences as seen in the visual world in general rather than the MSRC dataset. Given responses to these questions, for every category $z_i$, we count how often they preferred each category $z_j$ over the other categories. This gives us an estimate of $P(z_j|z_i)$ from humans. We compute $P(z_i)$ from the training images to obtain $P(z_i,z_j)$, which gives us a $21 \times 21$ co-occurrence matrix. We use the Chow-Liu algorithm on this matrix, as was used in~\cite{Jian} on the machine class co-occurrence potentials to obtain the tree structure, where the edges connect highly co-occurring nodes. As shown in Figure~\ref{fig:chowliu}, the structure of the human tree is quite similar to the tree obtained from the MSRC training set. For example, in both trees, there are edges between \emph{grass} and categories like \emph{cow}, \emph{sheep}, and \emph{flower}. However, some edges exist in the human tree that are missing in the machine tree e.g., the edge between sky and bird. 

\subsection{Detection}  

\noindent \textbf{Machine:} Detection is incorporated in the model by generating a large set of candidate bounding boxes using the deformable part-based model~\cite{s:felzenswalb10} which has multiple mixture components for each object class. The CRF model reasons about whether a detection is a false or true positive. On average, there are $16$ hypotheses per image. A binary variable  $b_i$ is  used for each detection and it is connected  to the binary class variable, $z_{c_i}$, where $c_i$ is the class of the detector that fired for the $i-$th hypothesis. 

\noindent \textbf{Human:} Since most objects in the MSRC dataset are quite big, it is expected that human object detection would be nearly perfect. As a crude proxy, we showed subjects images inside ground truth object bounding boxes and asked them to recognize the object. Performance  was almost perfect at 98.8\%. Hence, we use the ground truth object bounding boxes to simulate human responses. 
\begin{table*}[htp]
{\scriptsize
\begin{center}
\addtolength{\tabcolsep}{-0.55pt} 
\begin{tabular}{c|c|c|c|c|c|c|cc}
\cline{2-7}
&\multicolumn{3}{>{\columncolor{rr}}c|} {{\bf Oracle}} & \multicolumn{3}{>{\columncolor{orange}}c|}{{\bf Automatic}} & \\
\cline{2-9}
 & Detector &  Training Mask & Cluster & Detector & Dist. Tr. & Naive & \multicolumn{1}{c|}{{\bf Human}}& \multicolumn{1}{c|}{\bf GT}\\
 \hline
\multicolumn{1}{|c|}{Separate} & 78.5 & 88.4 & 75.3 & 72.7 & 78.8 & 74.3 & \multicolumn{1}{c|}{80.2} & \multicolumn{1}{c|}{93.1} \\
\hline
\hline
\multicolumn{1}{|c|}{Segmentation (Det.)}       & 77.7 & 78.4 & 77.5 & 77.2 & 76.8 & 76.3  & \multicolumn{1}{c|}{77.4}  & \multicolumn{1}{c|}{80.2}  \\
\hline
\multicolumn{1}{|c|}{Segmentation (GT)}         & 80.9 & 82.3 & 81.1 & 80.8 & 81.6 & 79.5  & \multicolumn{1}{c|}{80.8} & \multicolumn{1}{c|}{84.5} \\
\hline
\hline
\multicolumn{1}{|c|}{Object Det. (Det.)}        & 46.8 & 47.6 & 47.7 & 46.8 & 47.4 & 46.7  & \multicolumn{1}{c|}{47.2}  & \multicolumn{1}{c|}{48.5}  \\
\hline
\multicolumn{1}{|c|}{Object Det. (GT)}          & 95.8 & 90.3 & 95.1 & 93.9 & 93.3 & 94.5  & \multicolumn{1}{c|}{96.4} & \multicolumn{1}{c|}{92.7} \\
\hline
\end{tabular}
\end{center}
}
\caption{Accuracies of different shape priors inside and outside the model. Average recall and Average Precision is reported in the middle two rows and the bottom two rows, respectively.}
\label{table:shape}
\end{table*}
\subsection{Shape} 
\begin{figure}[tp]
\centering
\includegraphics[width=19pc]{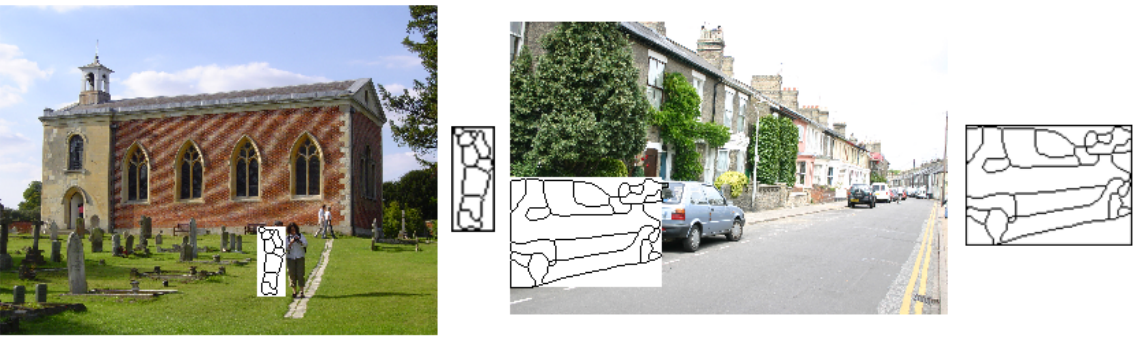}
\caption{Human object recognition from image boundaries. We show subjects segments inside the object bounding box and ask them to recognize the category of the object. We show the segments with (left image) and without (right image) context.}
\label{fig:boxtask}
\end{figure}
\noindent \textbf{Machine:} Shape potentials are incorporated in the model by connecting the binary detection variables $b_i$ to all segments $x_j$ inside the detection's bounding box. The prior is defined as an average training mask for each detector's mixture component. The values inside the mask represent the confidence that the corresponding pixel has the same label as the detector's class. In particular,  for the $i$-th candidate detection, this information is incorporated in the model by encouraging the $x_j$ segment to take class $c_i$ with strength proportional to the average mask values within the segment.

To evaluate the shape prior in isolation (outside the model), we quantify how well the ground truth mask matches the shape prior mask. The pixel-wise accuracy is normalized across foreground and background.  Accuracy by chance would be 50\%. The accuracy is computed only using pixels that fall in the bounding box. Table~\ref{table:shape} (top row) shows these results. The accuracy of the above approach ``Detector'' is 72.7\%. If an oracle were to pick the most accurate of the shapes (across the detector components), the accuracy would be 78.5\%\footnote{We asked humans to look at the average shape masks classify them into the object categories. Human performance at this task was 60\%.}.


We also experiment with alternative shape priors. We resize the binary object masks in the training images to $10\times10$ pixels. This produces a 100 dimensional vector for each mask. We use K-Means over these vectors to cluster the masks.  We set the number of clusters for each category to be equal to the number of detector's components for that category to be comparable to the detector prior. The shape mask for each cluster is the binary mask that is closest to the cluster center. If an oracle were to pick the most accurate of these K masks, it would achieve an accuracy of 75.3\% (``Cluster'' in Table~\ref{table:shape}), not better than the detector-based oracle above. If we were to pick the shape mask from the training images (without clustering) that matches the ground truth segmentation of an object the best, we get an oracle accuracy of 88.4\% (``Training Mask''). This gives us a sense of the accuracy one can hope to achieve by transferring shape masks from the training data without learning a generalization.


As another prior, we find the training mask whose shape matches the contours within a bounding box the best. We compute edges in the bounding boxes by thresholding the gPb contours. We compute the distance transform of these edges, and identify the training mask whose boundaries fall in regions closest to the edges. This training mask provides the shape prior for the bounding box. This \emph{automatic} approach (``Dist. Tr.'') has an accuracy of 78.8\% which is comparable to the \emph{oracle} on detector's masks.


Finally, we also experiment with a Naive approach. We simply encourage all segments that lie fully within the bounding box to take the corresponding class-label. The performance is 74.3\%, higher than automatically picking the detector mask using the mixture component!. This approach shares similarities with the superpixel straddling cue of \cite{objectness} which assumes that tight bounding boxes around objects do not have a lot of superpixels straddling the bounding box boundaries.

Note that if we ``snap'' the ground truth segmentation of an object to the segment boundaries i.e. each segment is turned on/off based on whether most of the pixels in the segment are foreground / background in the ground truth, we get an accuracy of 93.1\%. This is the upper-bound on the performance given the choice of segments.



\begin{figure}[tp]
\centering
\includegraphics[width=0.9\linewidth]{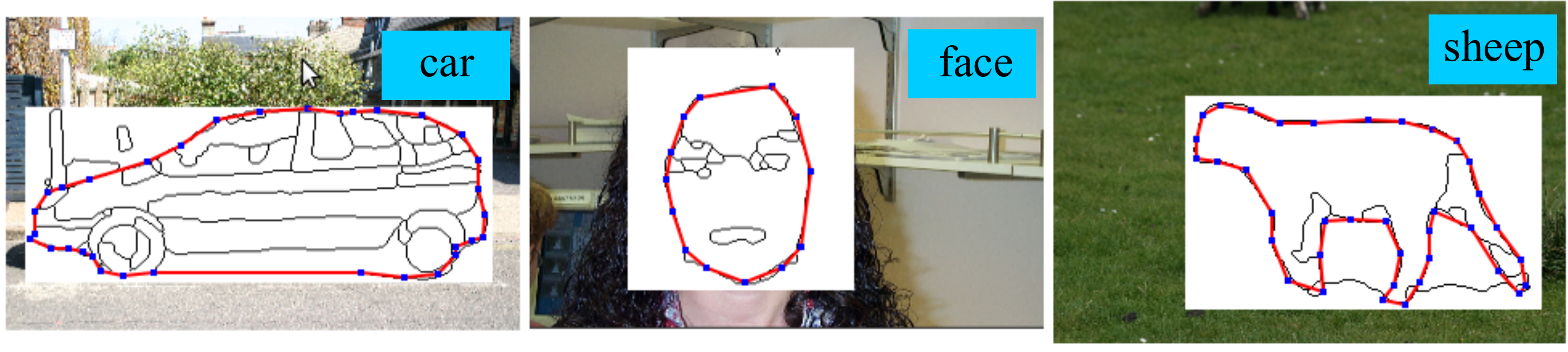}
\caption{Human shape mask labeling interface. Human subjects were asked to draw the object boundaries along the segment contours.}
\label{fig:sp_interface}
\end{figure}

\noindent \textbf{Human:} We showed 5 subjects the segment boundaries in the ground truth object bounding boxes along with its category label and contextual information from the rest of the scene. See  Figure~\ref{fig:sp_interface}. We showed subject contextual information around the bounding box because without it humans were unable to recognize the object category reliably using only the boundaries of the segments in the box (55\% accuracy). With context, classification accuracy was 94\%. See Figure~\ref{fig:boxtask} for example images. 


Using the interface of~\cite{Bharath11}, subjects were asked to trace a subset of the segment boundaries to match their expected shape of the object. The accuracy of the best of the 5 masks obtained for each object (normalized for foreground and background) was found to be 80.2\%. Recall that the best automatic accuracy we obtained with the machine was 78.8\% using the distance transform approach, not much worse than the human subjects' accuracy. This shows that humans can not decipher the shape of an object from the UCM segment boundaries much better than an automatic approach. Clearly, the UCM segment boundaries are not any more informative to humans than they are to machines.

\subsection{Scene and scene-class co-occurrence}  

\noindent \textbf{Machine:} We train a classifier~\cite{s:xiao10} to predict  each of the scene types, and use its confidence to form the unitary potential for the scene variable. The scene node connects to each binary class variable $z_i$ via a pairwise potential which is defined based on the co-occurrence statistics of the training data, i.e., likelihood of each class being present for each scene type.

\begin{figure}[tp]
\centering
\includegraphics[width=0.9\linewidth]{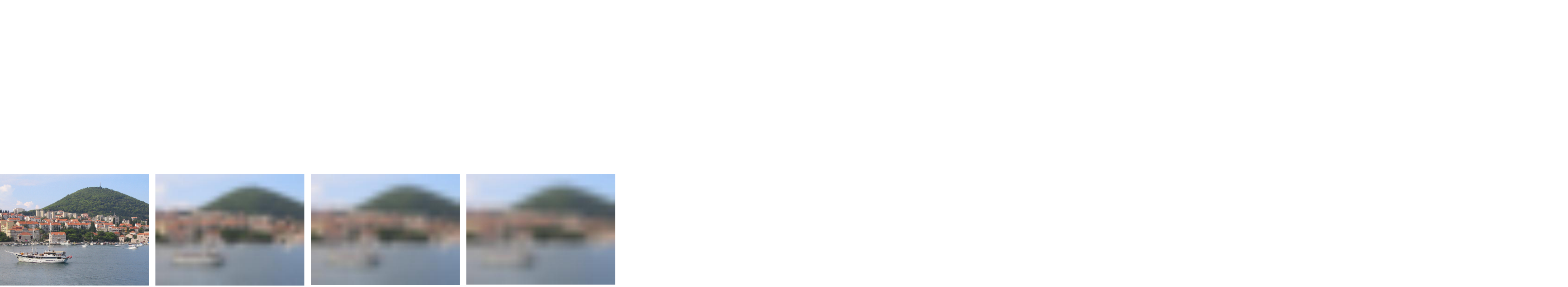}
\caption{Human scene classification. Subjects were shown images at multiple resolutions. Subjects were asked to choose the scene category for each image.}
\label{fig:scene}
\end{figure}

\noindent \textbf{Human:} To obtain scene unary, we ask human subjects to classify an image into one of the 21 scene categories used in~\cite{Jian} (see Table~\ref{tab:msrcinfo}). Images were presented at varying resolutions (i.e., original resolution,  smallest dimension rescaled to 32, 24 and 20 pixels) as shown in Figure \ref{fig:scene}. 
Subjects were allowed to select more than one category when confused, and the potential was computed as the proportion of responses each category got. Human accuracy at scene recognition was 90.4, 89.8, 86.8 and 85.3\% for the different resolutions, as compared to the machine accuracy of 81.8\%. Note that human performance is not 100\% even with full resolution images because the scene categories are semantically ambiguous. Humans clearly outperform the machine at scene recognition, but the question of interest is whether this will translate to improved performance for holistic scene understanding.

\begin{figure}[tp]
\centering
\includegraphics[width=\linewidth]{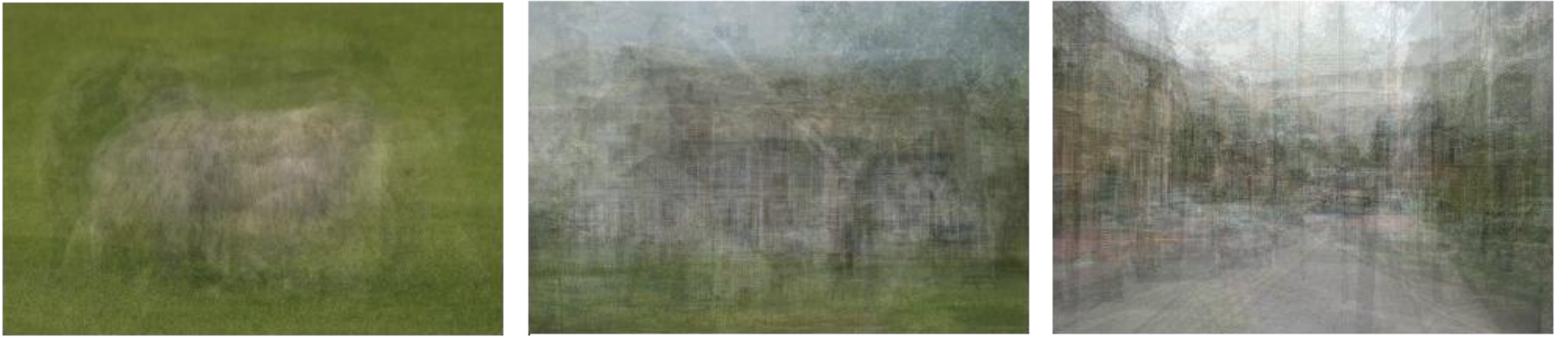}
\caption{Average scenes for some example scene categories in MSRC.}
\label{fig:avg_scene}
\end{figure}

Similar to the class-class experiment, to obtain scene-class co-occurrence statistics, subjects were asked which object category is more likely to be present in the scene. We ``show'' the scene either by naming its category (no visual information), or by showing them the average image for that scene category. Examples are shown in Figure \ref{fig:avg_scene}\footnote{When asked to look at the average images and recognize the scene category, subjects were 80\% accurate.}. The normalized co-occurrence matrix is then used as the pairwise potential.

\subsection{Ground truth Potentials} In addition to human potentials (which provide a lower-bound), we are also interested in establishing an upper-bound on the effect each subtask can have on segmentation performance by introducing ground truth (GT) potentials into the model. We formed each potential using the dataset annotations. For segments and super-segments we simply set the value of the potential to be $1$ for the segment GT label  and $0$ otherwise, similarly for  scene and class unary potentials. For object detection, we used the GT boxes as the candidates and set their detection scores to $1$. For the shape prior, we use a binary mask that indicates which pixels inside the GT object bounding box have the object's label.

Note that in theory, some other settings of the variables in the model might produce better results than using ground truth. Therefore, using the ground truth information for each sub-task might not result in a strict upper-bound. 
\section{Experiments with CRFs}
\label{sec:results}

We now describe the results of inserting the human potentials in the CRF model. We also investigated how plugging in GT potentials or discarding certain tasks all together affects performance on the MSRC dataset. For meaningful comparisons, CRF learning and inference is performed every time a potential is replaced, be it with {\bf (i) Human} or {\bf (ii) Machine} or {\bf (iii) GT} or {\bf (iv) Remove}. 


\label{sec:machineexp}


\begin{figure*}[tph]
 \centering
 \subfigure[Semantic Segmentation]{
 \includegraphics[width=18pc]{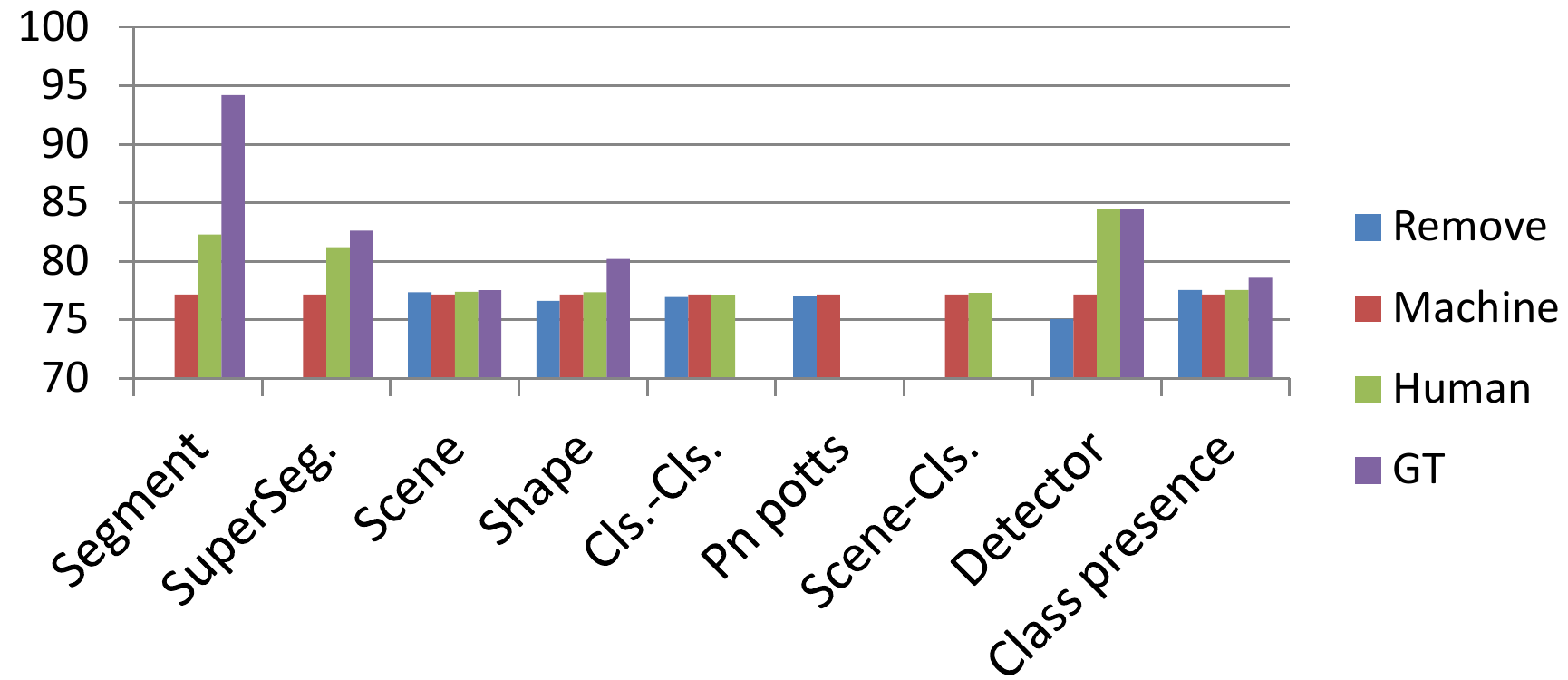}
 \label{fig:seg_pl}
 }
 \subfigure[Scene Classification]{
 \includegraphics[width=18pc]{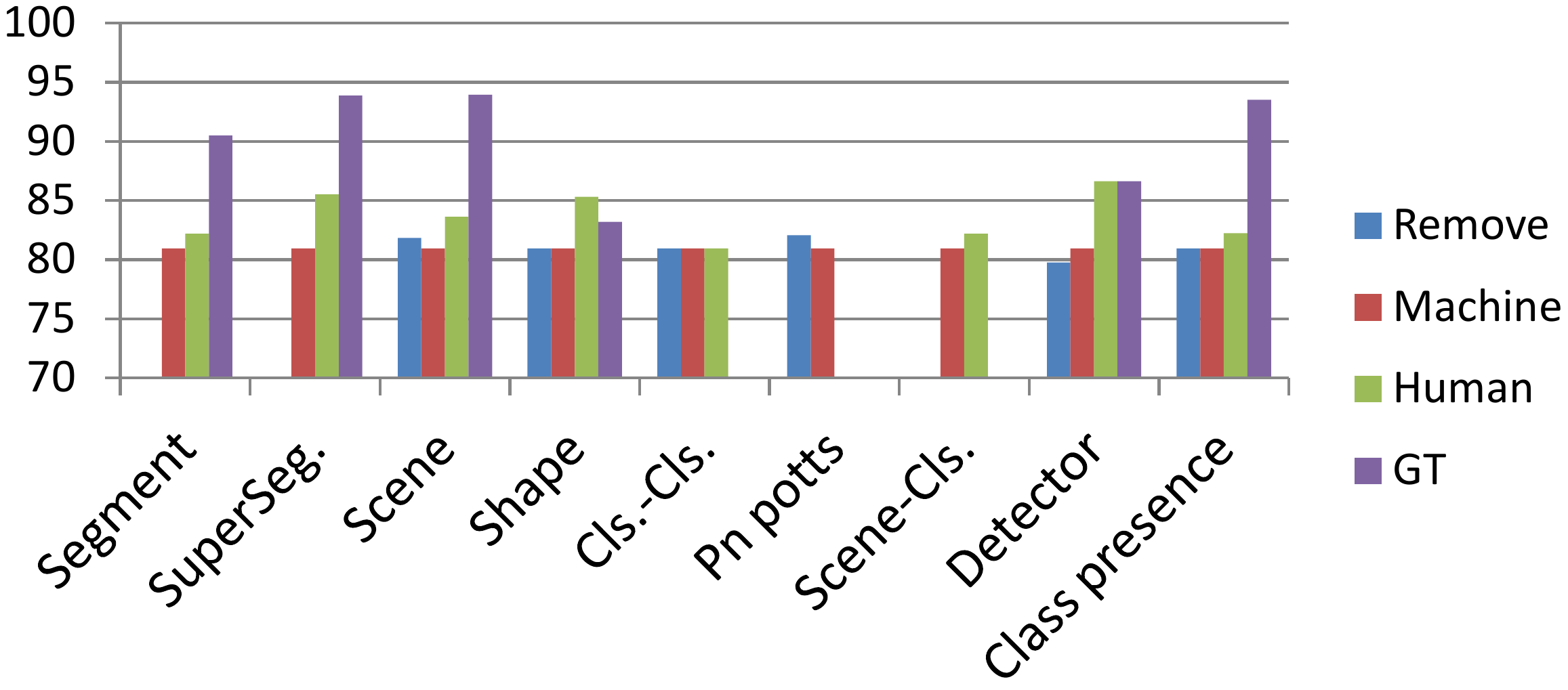}
 \label{fig:scene_pl}
 }
 \subfigure[Object Detection]{
 \includegraphics[width=18pc]{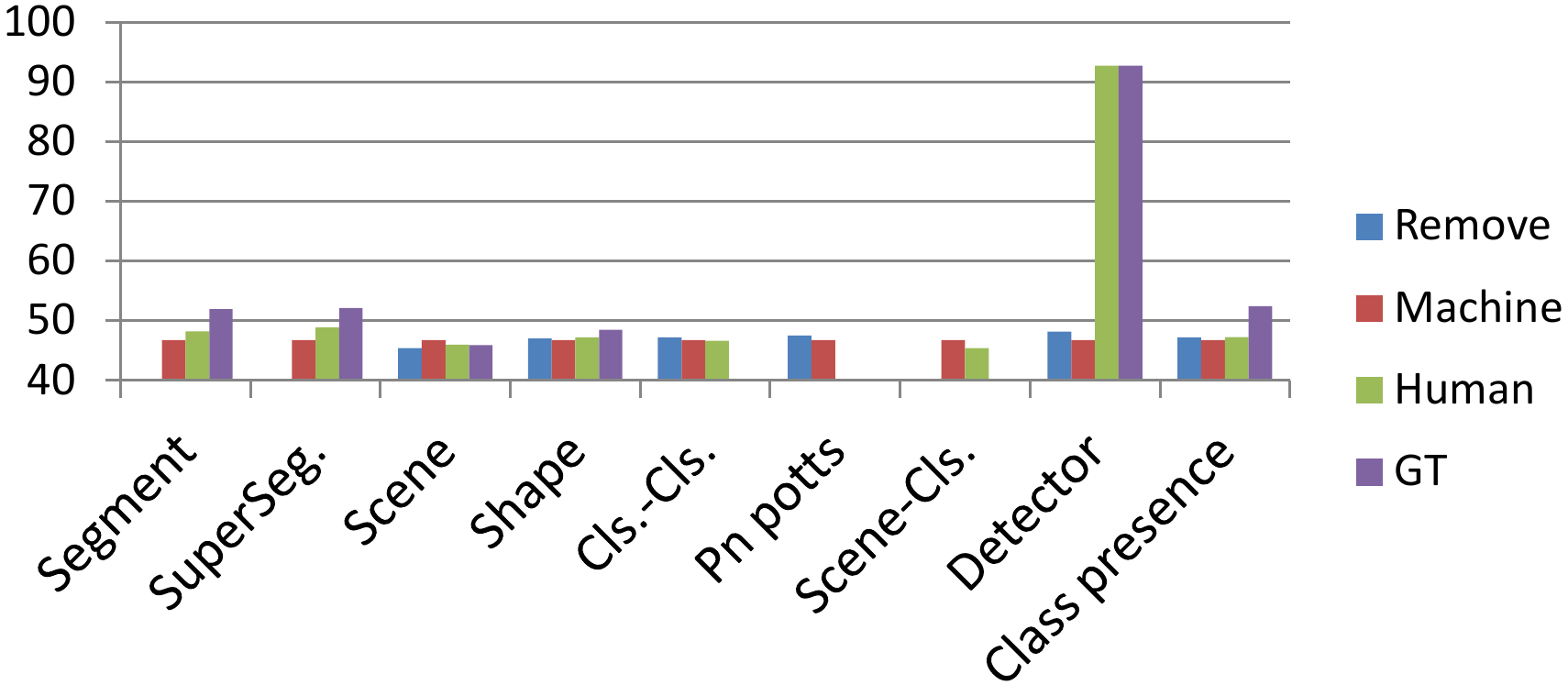}
 \label{fig:obj_pl}
 }
\caption{Impact of each component on machine holistic scene recognition. Here we show the semantic segmentation, object detection, and scene recognition accuracies when a single component of the model is changed (removed, implemented by a machine (default), replaced by a human or replaced with ground truth). The evaluation measures are average per-class recall, Average Precision (AP), and average recall for segmentation, object detection, and scene recognition respectively.}
\label{fig:sum_pl}
\end{figure*}

A summary of the results for the four different settings is shown in Figure~\ref{fig:sum_pl}.
Note that in each experiment only a \emph{single} machine potential was replaced, which is indicated in the x axis of the plots.  Missing bars for the \emph{remove} setting indicate that removing the corresponding potential would result in the CRF being disconnected, and hence that experiment was not performed. GT is not meaningful for pairwise potentials. The average over all categories is shown on the y axis. 

There are several interesting trends. Having GT information for class presence (i.e. knowing which objects are present in the image) clearly helps scene recognition, but also gives a noticeable boost to object detection and segmentation. This argues in favor of informative classifiers for class presence, which were not used in the current model~\cite{Jian}, but is, e.g., done in~\cite{s:gonfaus10}. Class-class co-occurrence potential and the scene-class potential have negligible impact on the performance of all three tasks. The choice of the scene classifier has little impact on the segmentation but influences detection accuracy. We find that human object detection boosts performance, which is not surprising. GT shape also improves performance, but as discussed earlier, we find that humans are unable to instantiate this potential using the UCM segment boundaries. This makes it unclear what the realizable potential of shape is for the MSRC dataset. One human potential that does improve performance is the unitary segment potential. This is quite striking since human labeling accuracy of segments was substantially worse than machine's (72.2\% vs. 77.4\%), but incorporating the potential in the model significantly boosts performance (from 77.2\% to 82.3\%).  Intrigued by this, we performed detailed analysis to identify properties of the human potential that are leading to this boost in performance. Resultant insights provided us concrete guidance to improve machine potentials and hence state-of-the-art accuracies.

\subsection{Analysis of segments in MSRC}
\label{sec:segmsrc}
We now describe the various hypotheses we explored including unsuccessful and successful ones to explain the boost provided by human segment potentials.

\noindent \textbf{Scale:} We noticed that the machine did not have access to the scale of the segments while humans did. So we added a feature that captured the size of a segment relative to the image and re-trained the unary machine potentials. The resultant segmentation accuracy of the CRF was 75.2\%, unfortunately worse than the original accuracy at 77.2\%.



\noindent \textbf{Over-fitting:} The machine segment unaries are trained on the same images as the CRF parameters, potentially leading to over-fitting. Humans obviously do not suffer from such biases. To alleviate any over-fitting in the machine model, we divided the training data into 10 partitions. We trained the machine unaries on 9 parts, and evaluated them on the $10^{th}$ part, repeating this 10 times. This gives us machine unaries on the entire training set, which can be used to train the CRF parameters. While the machine unaries may not be exactly calibrated, since the training splits are different by a small fraction of the images, we do not expect this to be a significant issue. The resultant accuracy was 76.5\%, again, not an improvement.


\noindent \textbf{Ranking of the correct label:} It is clear that the highest ranked label of the human potential is wrong more often than the highest ranked label of the machine potential (hence the lower accuracy of the former outside the model). But we wondered if perhaps even when wrong, the human potential gave a high enough score to the correct label making it revivable when used in the CRF, while the machine was more ``blatantly'' wrong. We found that among the misclassified segments, the rank of the correct label using human potentials was 4.59 -- better than 6.19 (out of 21) by the machine. 

\noindent \textbf{Uniform potentials for small segments:} Recall that we did not have human subjects label the segments smaller than 500 pixels and assigned a uniform potential to those segments. The machine on the other hand produced a potential for each segment. We suspected that ignoring the small (likely to be misclassified) segments may give the human potential an advantage in the model. So we replaced the machine potentials for small segments with a uniform distribution over the categories. The average accuracy unfortunately dropped to 76.5\%. As a follow-up, we also weighted the machine potentials by the size of the corresponding segment. The segmentation accuracy was still 77.1\%, similar to the original 77.2\%.


\noindent \textbf{Regressing to human potentials:} We then attempted to directly regress from the machine potential as well as the segment features (TextonBoost, LBP, SIFT, ColorSIFT, location and scale) to the human potential, with the hope that if for each segment, we can predict the human potential, we may be able to reproduce the high performance. We used Gaussian Process regression with an RBF kernel. The average accuracy in both cases was lower: 75.6\% and 76.5\%. We also replicated the sparsity of human potentials in the machine potentials, but this did not improve performance by much (77.3\%). 

\begin{figure}[tp]
\centering
\includegraphics[width=15pc]{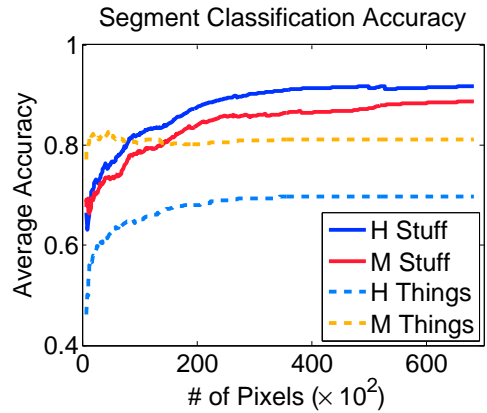}
\caption{Humans (H) and machines (M) have different performance for recognizing stuff and things segments. Humans are generally better at recognizing stuff, while machines are better at things recognition. Larger segments are generally easier to recognize.}
\label{fig:th_vs_stuff}
\end{figure}

\noindent \textbf{Complementarity:} To get a deeper understanding as to why human segment potentials significantly increase performance when used in the model, we performed a variety of additional hybrid CRF experiments. These included having human (H) or machine (M) potentials for segments (S) or super-segments (SS) or both, with or without the $P^n$ potential in the model. The results are shown in Table~\ref{table:pn}. The last two rows correspond to the case where both human and machine segment potentials are used together at the same level. In this case, using a $P^n$ potential or not has little impact on the accuracy. But when the human and machine potentials are placed at different levels in the model (rows 3 and 4), not having a $P^n$ potential (and thus loosing connection between the two levels) significantly hurts performance. This indicates that even though human potentials are not significantly more accurate than machine potentials, when both human and machine potentials interact, there is a significant boost in performance, demonstrating the complementary nature of the two. 


\begin{table}[h]
{\scriptsize
\begin{center}
\begin{tabular}{|c|c|c|c|c|}

\hline
&$P^n$&without $P^n$\\
\hline
H S, H SS      & 78.9 & 77.2  \\
M S, M SS      & 77.2 & 77.0\\
H S, M SS      & 82.3 & 75.3\\
M S, H SS      & 81.2 & 78.2 \\
\rowcolor{yellow} H S+M S, M SS & 80.9 & 81.3 \\
\rowcolor{yellow} H S+M S, H SS & 82.3 & 82.8 \\
\hline
\end{tabular}
\end{center}
}
\caption{Human and machine segment potentials are complementary. The last two rows correspond to the case where both human and machine segment potentials are used together at the same level. In this case, using a $P^n$ potential or not has little impact on the accuracy. But when the human and machine potentials are placed at different levels in the model (rows 3 and 4), not having a $P^n$ potential (and thus loosing connection between the two levels) significantly hurts performance. This indicates that even though human potentials are not significantly more accurate than machine potentials, when both human and machine potentials interact, there is a significant boost in performance, demonstrating the complimentary nature of the two.}
\label{table:pn}
\end{table}



\begin{figure*}[tph]
 \centering
 \subfigure[Machine classification with 200x200 windows]{
 \includegraphics[width=15pc]{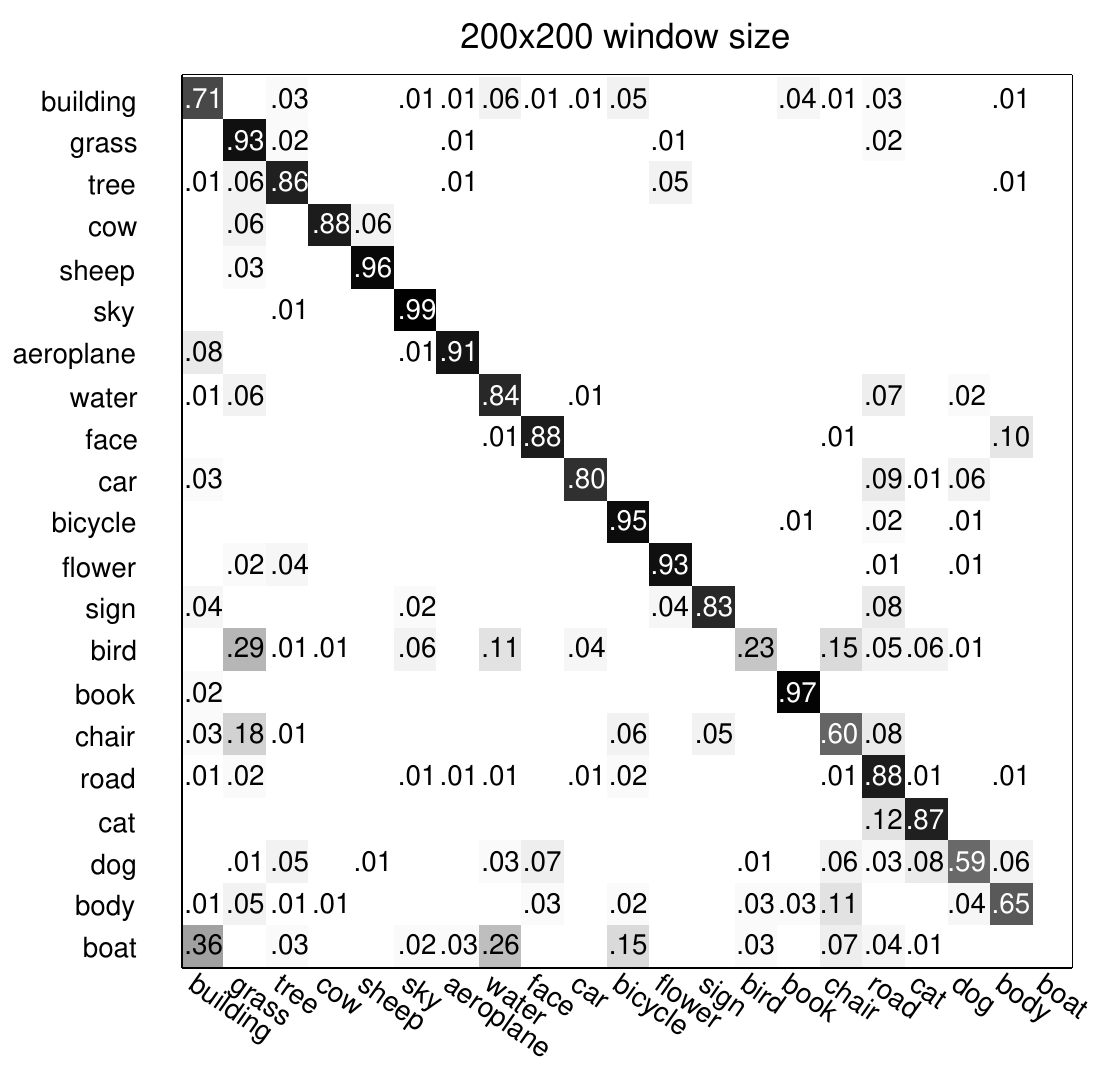}
 \label{fig:cm_mach}
 }
 \subfigure[Human classification]{
 \includegraphics[width=15pc]{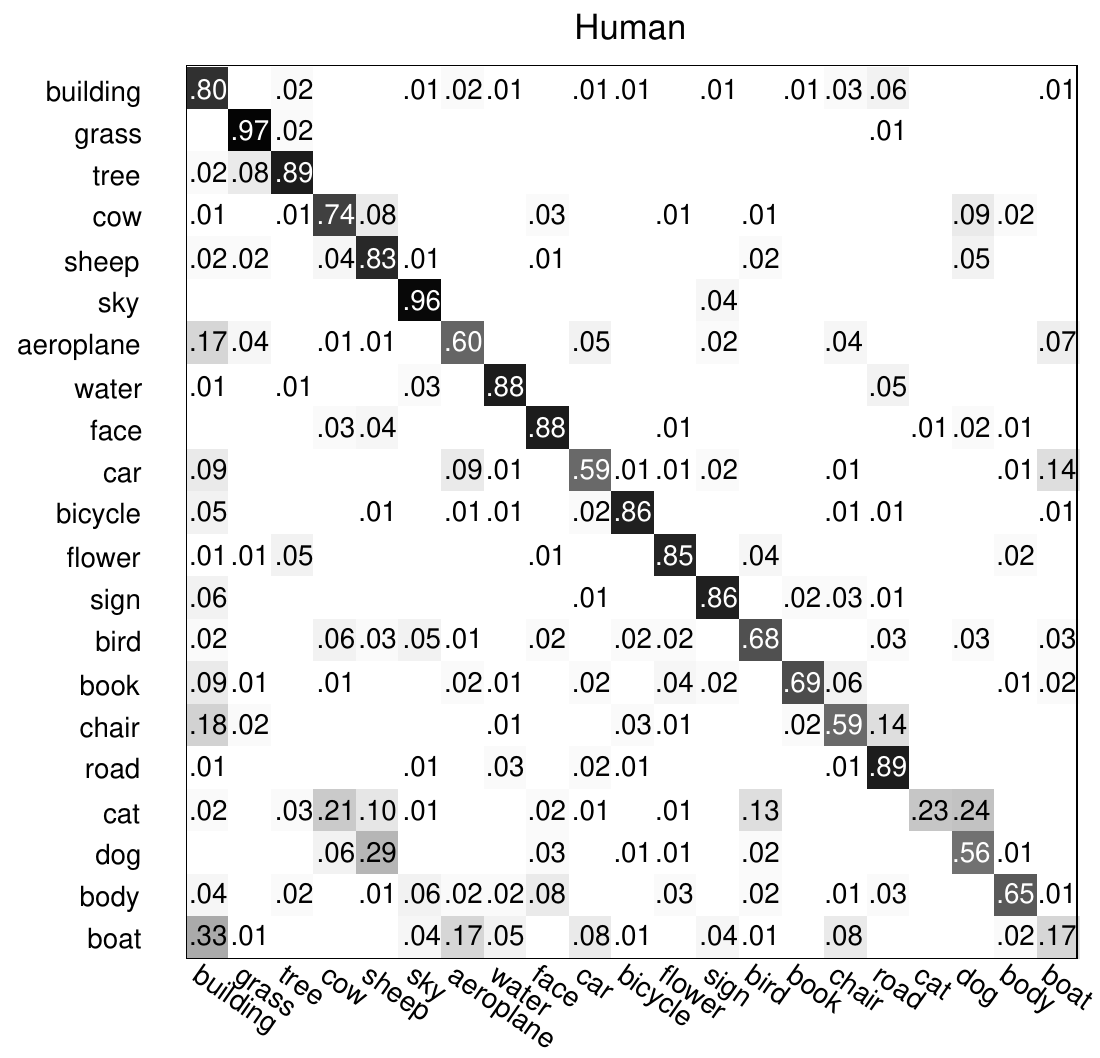}
 \label{fig:cm_hum}
 }
 \subfigure[Machine classification with 30x30 windows]{
 \includegraphics[width=15pc]{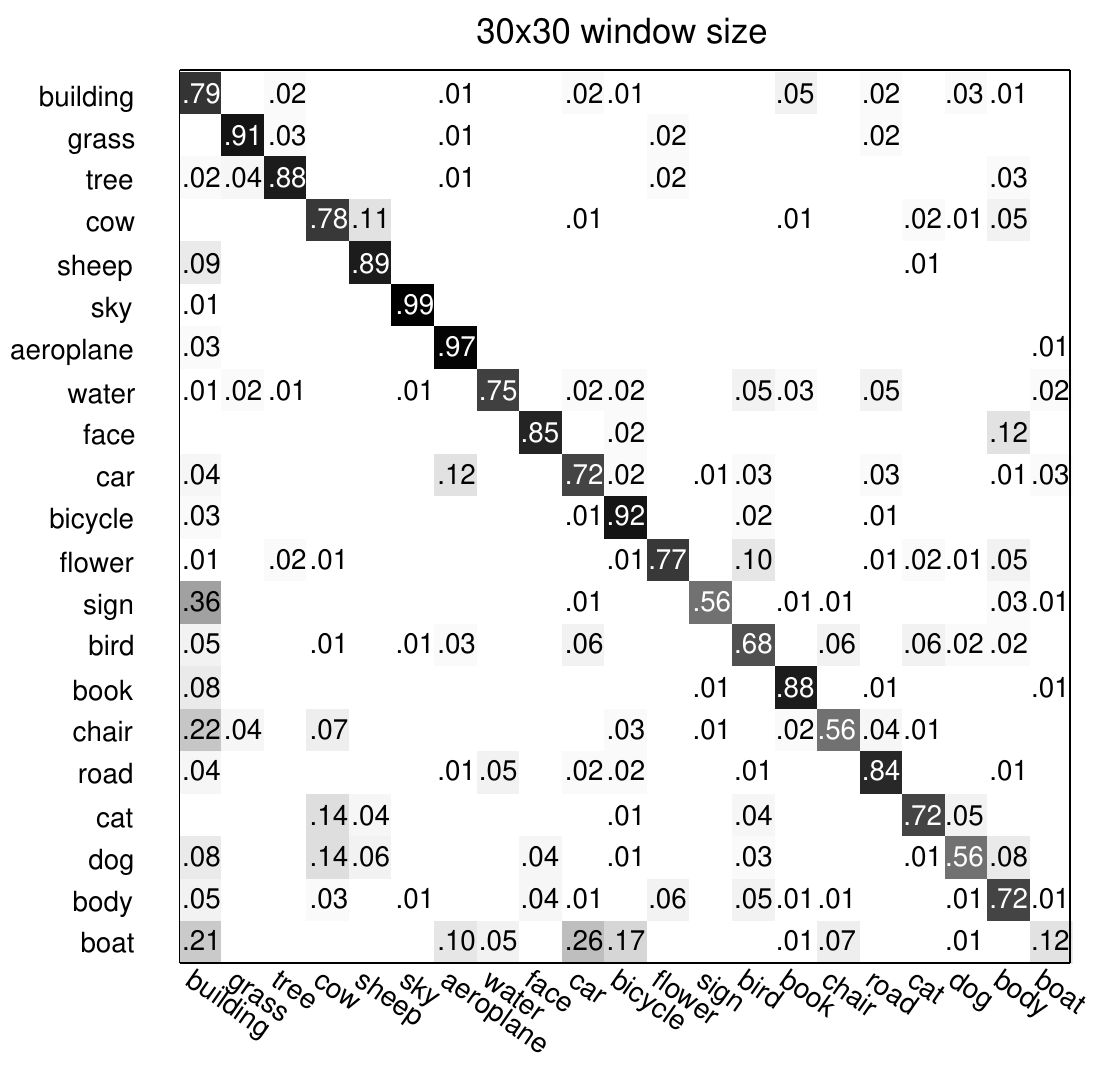}
 \label{fig:cm_small}
 }

\caption{The confusion matrices for segment classification are shown. For machine with large window size (a), there is high confusion between classes appearing in the surrounding area of each other, for example, bird-grass, car-road, etc. The types of mistakes are different for humans (b). They confused objects that look similar, for instance, there is confusion between cat-cow or boat-aeroplane. When we reduce the window size for machine (c), the mistakes become more similar to the human mistakes. Combining the small-window machine potentials with the large-window machine potentials results in a significant improvement in segmentation accuracy, resulting in state-of-the-art performance on the MSRC dataset.}
\label{fig:confusion}
\end{figure*}

Therefore, we hypothesized that the types of mistakes that the machine and humans make may be different. Our initial analysis showed that humans are generally better at detecting stuff while machine is better recognizing things (Figure~\ref{fig:th_vs_stuff}).

Additionally, we qualitatively analyzed the confusion matrices for both (Figure~\ref{fig:confusion}). We noticed that the machine confuses categories that spatially surround each other \eg bird and grass or car and road. This was also observed in~\cite{s:shotton09}. This is understandable because Textonboost uses a large (200 $\times$ 200) window surrounding a pixel to generate its feature descriptor. On the other hand, human mistakes are between visually similar categories \eg car and boat.\footnote{One consequence of this is that the mistakes made within a super-segment are consistent for machines but variable for humans. Specifically, on average machine assigns different labels to 4.9\% of segments, while humans assign different labels to 12\% of the segments within a super-segment. The consistent mistakes may be harder for other components in the CRF to fix.} Hence, we trained Textonboost with smaller windows. We re-computed the segment unaries and plugged them into the model in addition to the original unaries that used large windows. The average accuracy using window sizes of 10, 20, 30 and 40 are shown in Table~\ref{table:finalres}. The accuracy outside the model is shown in Table~\ref{table:outseg}. This improvement of 2.4\% over state-of-the-art is quite significant for this dataset\footnote{Adding a new unary potential simply by incorporating a different set of features and kernels than Textonboost (such as color, SIFT and self-similarity with intersection kernel) provides only a small boost at best ($77.9\%$).}. Notice that the improvement provided by the entire CRF model over the original machine unaries \emph{alone} was 3\% (from 74.2\% to 77.2\%). While a fairly straightforward change in the training of machine unaries lead to this improvement in performance, we note that the insight to do so was provided by our use of humans to ``debug'' the state-of-the-art model.

\begin{table}[h]
\setlength{\tabcolsep}{3pt}
{\scriptsize
\begin{center}
\begin{tabular}{|c|c|c|c|c|c|}
\hline
\shortstack{Textonboost \\ Window Size} & \cite{Jian} (200$\times$200) & 10$\times$10 & 20$\times$20 & 30$\times$30 &  40$\times$40 \\
\hline
Average per-class Recall & 77.2 & 77.9 & 78.5 & 79.6 & 79.6 \\
\hline
\end{tabular}
\end{center}
}
\caption{Segmentation accuracy obtained by resizing Textonboost window size. Decreasing the window size makes the machine errors similar to human errors in the task of isolated segment classification.}
\label{table:finalres}
\end{table}

\begin{table}[h]
\setlength{\tabcolsep}{3pt}
{\scriptsize
\begin{center}
\begin{tabular}{|c|c|c|c|c|c|c|}
\hline
\shortstack{Textonboost \\ Window Size} & 200$\times$200 & 10$\times$10 & 20$\times$20 & 30$\times$30 &  40$\times$40 & Humans\\
\hline
\shortstack{Average per-class Recall \\ outside the model} & 77.2 & 66.3 & 70.8 & 75.6 & 77.2 & 72.2 \\
\hline
\end{tabular}
\end{center}
}
\caption{Segmentation accuracy obtained by humans and by resizing Textonboost window size outside the model. The accuracy corresponds to segments larger than 500 pixels.}
\label{table:outseg}
\end{table}

\subsection{Analysis of Segment Classification in PASCAL}
PASCAL is labeled only with 20 ``things", making it uninteresting for holistic scene understanding. Hence, we used annotations that include 14 additional labels shown in Table \ref{table:pasc_cls} for PASCAL 2010 dataset. These 14 classes appear frequently at the immediate surrounding of the original 20 object classes. We used 651 images for training the machine classifier and tested the learned classifier on 200 random images. We chose the random subset such that it has a similar pixel-wise class distribution to the full dataset. We used the same MTurk setup as Figure~\ref{fig:interf} to carry out the human experiments on these 200 images.

\begin{table}[!ht]

\addtolength{\tabcolsep}{-0.1cm}
\begin{center}
\begin{tabular}{|c|c|c|c|c|c|c|}
\multicolumn{7}{>{\columncolor{blb}}c}{{\bf Additional PASCAL classes}} \\
\hline
sky & grass & ground & road & building & tree & water \\
\hline
mountain & wall & floor & railroad & keyboard & door & ceiling \\
\hline
\end{tabular}
\end{center}
\caption{We added annotations for the above semantic classes to PASCAL dataset.}
\label{table:pasc_cls}
\end{table}

For classification, we used the method of \cite{Carreira12}. The only difference is that we use patches of different sizes as the input to the classifier instead of their CPMC segments. We use square patches to be consistent with TextonBoost that was used for MSRC. The reason that we did not use TextonBoost for this experiment is that it does not scale well to larger number of images and categories. The patches are centered at the center of superpixels so there is a patch associated to each superpixel. We used 30x30 and 100x100 patches for our experiments. The confusion matrices are shown in the supplementary document due to space limitation. 

A similar pattern of confusion exists in the PASCAL dataset. For example, for humans, there is confusion between aeroplane and semantically similar categories such as car and train, while the machine confuses aeroplane with sky or road that appear at immediate surroundings of aeroplanes and the confusion with car or train is negligible. Therefore, the types of machine and human mistakes are different for PASCAL dataset as well. Also, similar to the MSRC case, as we decrease the patch size, the machine error becomes more similar to the human case. For instance, there is 31\% confusion between boat and building when we use $100\times 100$ patches, but when we reduce the patch size, the confusion becomes 51\% (closer to 62\% for humans). To quantify this, we computed the symmetric Kullback-Leibler (KL) distance between the corresponding rows of the confusion matricies and summed over the distances between the rows. The distance between Human and Machine $100\times 100$ is 294.35 while it is 256.25 between Human and Machine $30\times 30$. We see that the Machine 30x30 makes more human-like mistakes than Machine 100x100. Similarly, for MSRC, the distance between human and machine with large window was larger than human and machine with small window (151.2 vs. 63.8).

The accuracy for Machine with $100\times 100$ patches is 28.1\% and for human is 50.7\%. To show that human and machine mistakes are complementary, we used an oracle that picked the segment label provided by humans if machine made a mistake. The pixel-wise accuracy of this approach is 56.5\%, which is higher than both human and machine accuracies and shows the complementary nature of the mistakes. Designing machine potentials that make human-like mistakes (even if they are not as accurate as humans) may improve machine performance, as we demonstrated on the MSRC dataset in Section~\ref{sec:segmsrc}. Note that these experiments are independent of the system in \cite{Jian}.

%

\subsection{Analysis of Shape Prior}
In section~\ref{sec:pots}, we explored various types of shape priors outside the machine model. We now evaluate the impact of the various shape potentials when used in the full model. The second two rows in Table~\ref{table:shape} provide the semantic segmentation accuracy, while bottom two rows correspond to object detection accuracy. The bottom row in both pairs of rows corresponds to using the ground truth bounding boxes around an object, while the top rows correspond to using an object detector to automatically determine the object bounding boxes.

The improvement in performance we obtained over the detector induced prior used in~\cite{Jian} via our distance transform over contour detection approach outside the model (Table~\ref{table:shape} top row) did not translate into an improvement in the performance of the overall model. Further analysis into this revealed that while the normalized binary segmentation accuracy for our approach was better, the unnormalized accuracy was slightly worse, which corresponds better to the metric the model is trained to optimize. While GT shape can provide significant improvement in conjunction with GT bounding boxes for objects, we find that human subjects were not able to realize this potential in terms of segmentation accuracy. Interestingly, for object detection, using human shape on ground truth detection outperforms using ground truth shape on ground truth detection. 



%

\section{Analyzing the pipeline} 
\label{sec:modeljust}

In this section, we analyze the holistic scene understanding CRF model using the MSRC dataset. First, we investigate whether the components used in the model are beneficial for humans. Second, we estimate the potential that the model as a whole holds for all the three tasks.

\subsection{Contextual Image Labeling}

\begin{figure*}[tp]
\centering
\begin{minipage}{1\linewidth}
\includegraphics[width=1\linewidth,trim=0 00 0 0,clip=true]{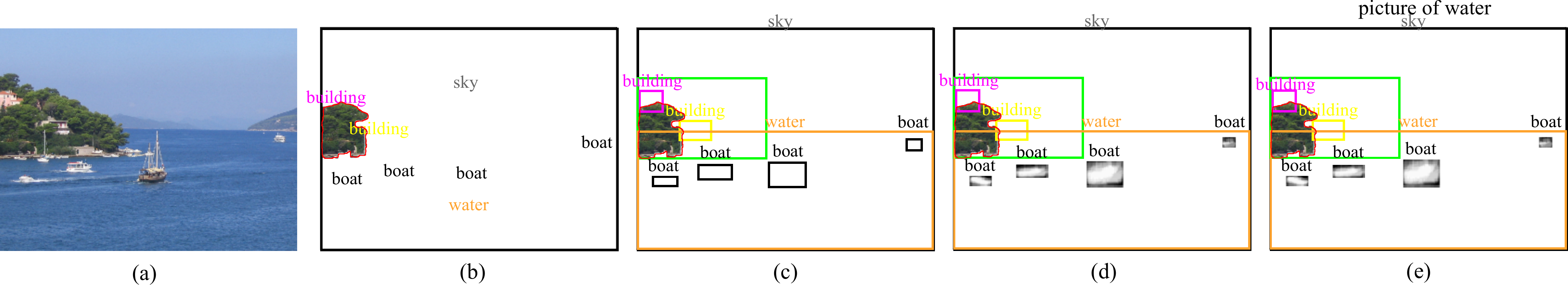}
\caption{(b):(e) Human-study interfaces for (a) with increasingly more contextual information.} 
\label{fig:context}
\end{minipage}
\end{figure*}

To study if humans benefit from contextual information provided by the model in order to recognize local regions, we design an experiment where, in addition to the segment pixels, we show subjects progressively more information from the model: presence / absence of each class, object bounding boxes, shape prior masks and scene type. We selected 50 random segments of any size from each category and asked the subjects to classify each segment. Figure~\ref{fig:context} shows an example for each of the interfaces used for this study. From left to right (b -- e), the contextual information available to the subjects is increased. 

The results are shown in Table \ref{tab:resultsClean}. We see that human performance in the task of image labeling significantly increases when presented with the same contextual information used by holistic machine models. We also show accuracies of the machine model when given access to precisely the same information as shown to humans. We find that humans can effectively leverage the available information. Even though human performance is worse than machines when viewing segments in isolation, they outperform the machine when given access to the contextual information modeled by the CRF. This is especially true for ``things''. Access to context seem to confuse human responses for ``stuff''. Investigating this further is part of future work. 

\begin{table*}[!ht]
{\scriptsize
\addtolength{\tabcolsep}{-0.1cm}
\begin{center}
\begin{tabular}{l|ccccccccccccccccccccc|ccc}
 & \rotatebox{90}{build.} & \rotatebox{90}{grass} & \rotatebox{90}{tree} & \rotatebox{90}{cow} & \rotatebox{90}{sheep} & \rotatebox{90}{sky} & \rotatebox{90}{aeropl.} & \rotatebox{90}{water} & \rotatebox{90}{face} & \rotatebox{90}{car} & \rotatebox{90}{bicycle} & \rotatebox{90}{flower} & \rotatebox{90}{sign} & \rotatebox{90}{bird} & \rotatebox{90}{book} & \rotatebox{90}{chair} & \rotatebox{90}{road} & \rotatebox{90}{cat} & \rotatebox{90}{dog} & \rotatebox{90}{body} & \rotatebox{90}{boat} & \rotatebox{90}{{\bf avg.}} & \rotatebox{90}{{\bf avg. stuff}} & \rotatebox{90}{{\bf avg. things}}\\ 
\hline
\hline
& \multicolumn{24}{>{\columncolor{blb}}c|}{{\bf Humans}} \\
\hline
Segm. (S)   & 48 & 82 & 38 & 20 & 14 & {\bf 64} & 18 & 60 & 50 & 22 & 12 & 58 & 44 & 8 & 20 & 16 & 36 & 0 & 6 & 22 & 6 & 30.7 & 54.7 & 21.1 \\
S+label  & {\bf 78} & {\bf 84}   & {\bf 54} & 86 & 82 & 58 & 64 & 62 & 72 & 50 & 72 & 90 & 92 & 88 & 90 & 90 & {\bf 52} & 92 & 82 & 58 & 70 &  74.6 & {\bf 64.7} & 78.5\\
S+box (B)           & 58 & 72           & 38 & {\bf 100} & 98 & 44 & {\bf 100} & 54 & 82 & 96 & {\bf 94} & {\bf 100} & {\bf 96} & {\bf 96} & {\bf 98} & 96 & 46 & {\bf 100} & {\bf 98} & 92 & 98 & 83.6 & 52.0 & 96.3 \\
S+B+msk & 58  & 72 & 44 & {\bf 100} & {\bf 100} & 42 & {\bf 100} & 54 & 98 & {\bf 98} & {\bf 94} & {\bf 100} & {\bf 96} & 94 & {\bf 98} & {\bf 98} & 36 & {\bf 100} & {\bf 98} & {\bf 96} & {\bf 100}  & 84.6 & 51.0 & {\bf 98.0} \\
Full info.            & 68          & 78            & 48 &  98 & {\bf 100} & 50 & {\bf 100} & {\bf 66} & {\bf 100} & {\bf 98} & {\bf 94} & {\bf 100} & {\bf 96} & 94 & {\bf 98} & {\bf 98} & 46 & {\bf 100} & {\bf 98} & 90 & {\bf 100} & {\bf 86.7} & 59.3 & 97.6 \\
\hline
& \multicolumn{24}{>{\columncolor{rr}}c|}{{\bf Machine model}}  \\
\hline
Segm. (S)   & 82 & 86 & 93 & 74 & 94 & 96 & 84 & 88 & 96 & 70 & 90  & 88  & 80 & 6 & 97   & 30 & 92 & 95 & 51 & 34 & 0 & 72.7 & 89.5 & 65.9\\
S+label        & 83 & 87 & 93 & 76 & 92 & 97 & 87 & 92 & 96 & 73 & 100 & 98  & 80 & 42 & 97  & 54 & 93 & 96 & 70 & 33 & 0 & 78.2 & 90.8 & 72.9\\
S+box (B)          & 83 & 87 & 93 & 86 & 96 & 97 & 90 & 92 & 96 & 82 & 100 & 100 & 87 & 44 & 100 & 61 & 93 & 96 & 71 & 35 & 55 & 82.9  & 90.8 & 79.9\\
S+B+msk     & 84 & 87 & 93 & 86 & 98 & 97 & 90 & 92 & {\bf 97} & 82 & 100 & 100 & 87 & 44 & 100 & 61 & 93 & 96 & 71 & 42 & 75  & 84.3 & {\bf 91.0} & {\bf 81.9} \\
Full info.     & \bf{84} & \bf{87} & \bf{93} & \bf{86} & \bf{98} & \bf{97} & \bf{90} & \bf{92} & 96 & \bf{82} & \bf{100} & \bf{100} & \bf{87} & \bf{44} & \bf{100} & \bf{61} & \bf{93} & \bf{96} & \bf{71} & \bf{42} & \bf{75} & {\bf 84.4} & {\bf 91.0} & {\bf 81.9} \\
\hline
\end{tabular}
\end{center}
}
\caption{Human segmentation accuracies with increasing information from the model}
\label{tab:resultsClean}
\end{table*}

\subsection{Journey to Perfection}
\begin{figure}[h]
\centering
\includegraphics[width=20pc]{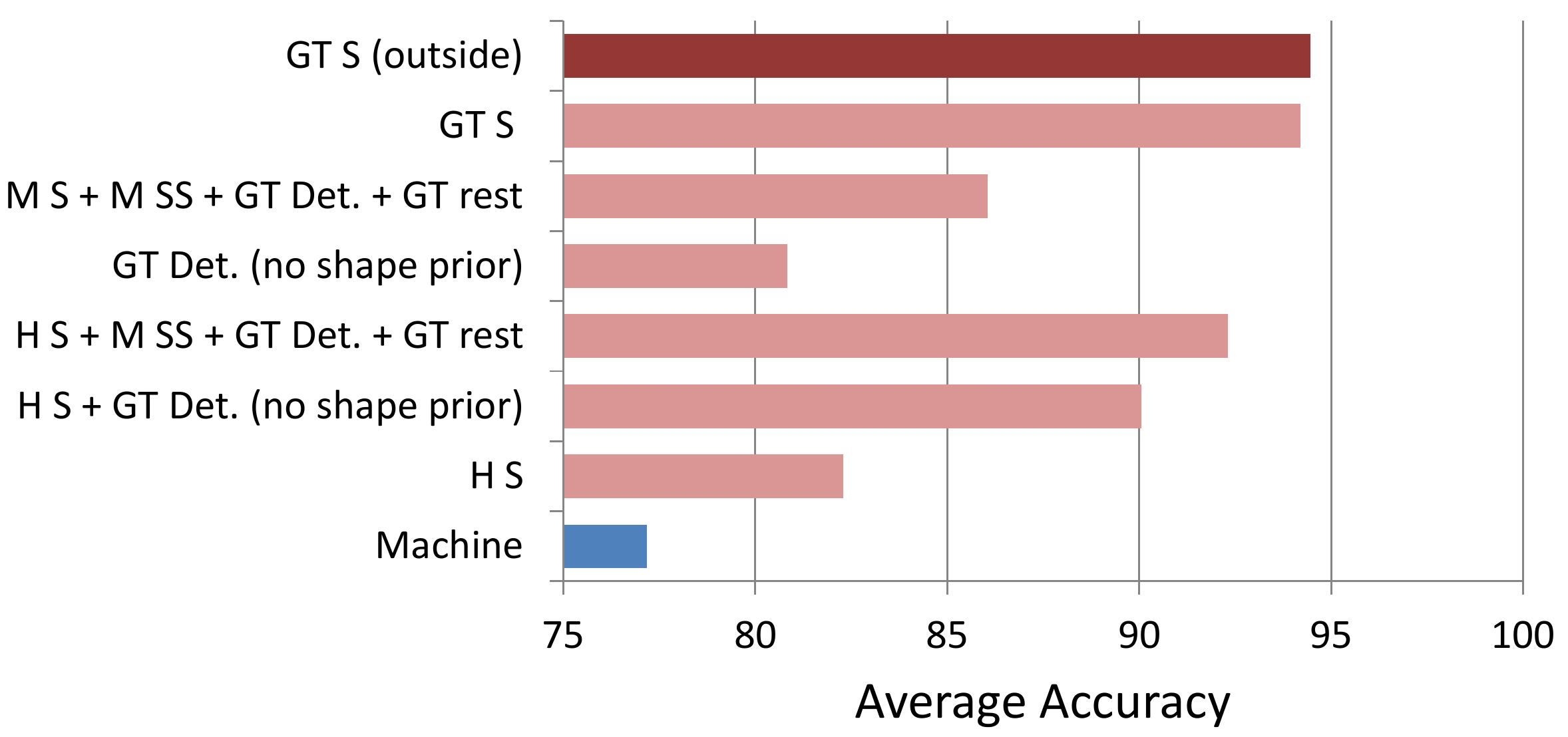}
\caption{Journey to perfection for segmentation.}
\label{fig:seg_jour}
\end{figure}

To analyze if the model has the potential to reach perfection, we conduct experiments using different combinations of machine, human, and ground truth potentials. Figure~\ref{fig:seg_jour} provides a journey on the segmentation task from the machine's current 77\% performance to perfection. We find that incorporating human (H) segment (S) potentials improves performance by 5-6\%. Incorporating ground truth (GT) detection (Det) provides an improvement of 4-7\%. Adding in GT for remaining tasks (except super-segments (SS)) further improves performance by 2-5\%. These combined bring us to 92\%. GT segments plugged into the model perform at 94.2\%, which outside the model are at 94.5\% (the upper-bound on the performance of the model since it makes segment-level decisions). This shows that as far as segmentation goes the model itself is sufficient and all the required tasks are being modeled. This analysis also provides concrete guidance for future research efforts: designing representations complementary to the ones used in the model, perhaps by mimicking human responses, has potential for significant improvement. And of course, improving on all the tasks would lead to more effective holistic scene understanding. Note that, holistically accurate systems do not require extremely high performance on each sub-task to achieve high performance.

\begin{figure}[h]
\centering
\includegraphics[width=20pc]{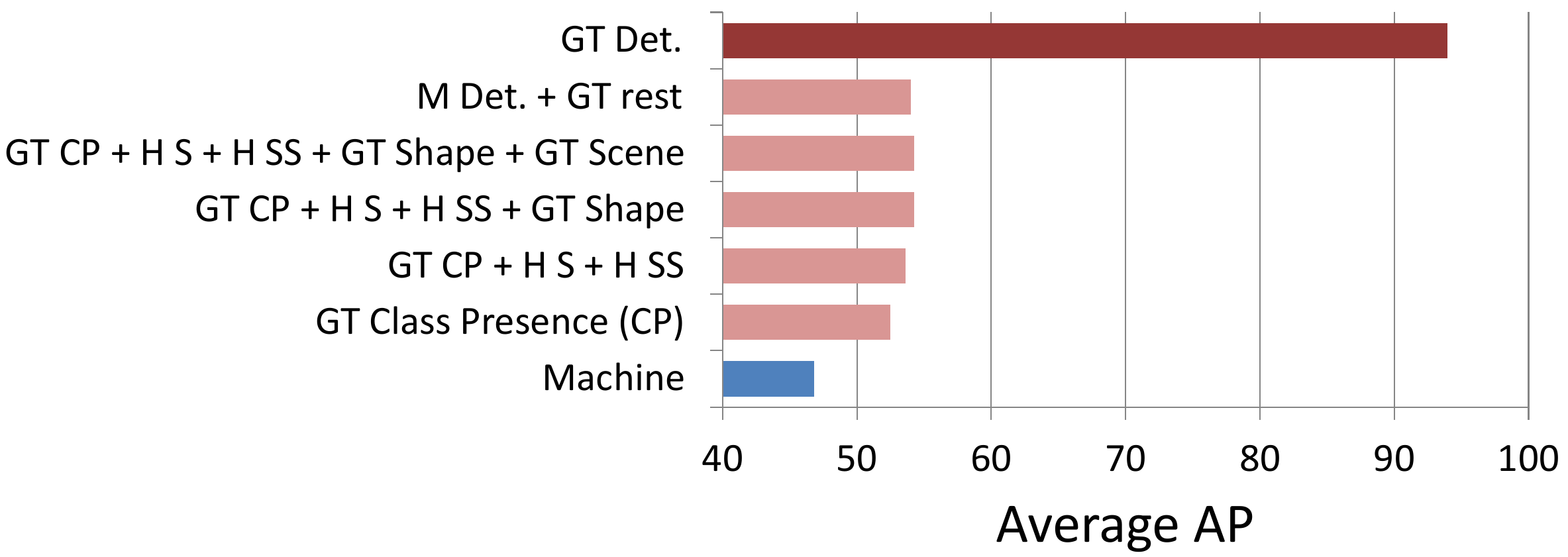}
\caption{Journey to perfection for object detection.}
\label{fig:det_jour}
\end{figure}

Figure~\ref{fig:det_jour} shows the effect of each component on the object detection task. The machine's average precision is 46.8. Including the ground truth information for class presence (CP) improves the average precision to 52.5 (5.7 improvement in AP). Incorporating human segment (S) and super-segment (SS) potentials provides an additional improvement of 1.1 AP, which leads to 53.6 AP. Ground truth shape information also provides improvement for the object detection task, and increases the AP to 54.3. Including the ground truth scene has negligible impact on performance. If we replace the human segment and super-segment potentials by their ground truth counterparts, the average precision decreases to 54.0. Hence, ground truth class-presence (knowing whether a certain category exists in the image or not similar to the image classification task in PASCAL) is the single component that provides the biggest boost in performance. Obviously, using ground truth information for object detection has a significant effect on the performance, where it improves the AP to 93.9. Hence, given the scope of this model, the burden of improving the detection performance from 54.3 to 93.9 lies on the detector itself. Enhancing the model with additional cues such as a rough 3D layout of the scene, etc. that directly influence likely locations and scales of detections may be important to aid the detector.

\begin{figure}[h]
\centering
\includegraphics[width=20pc]{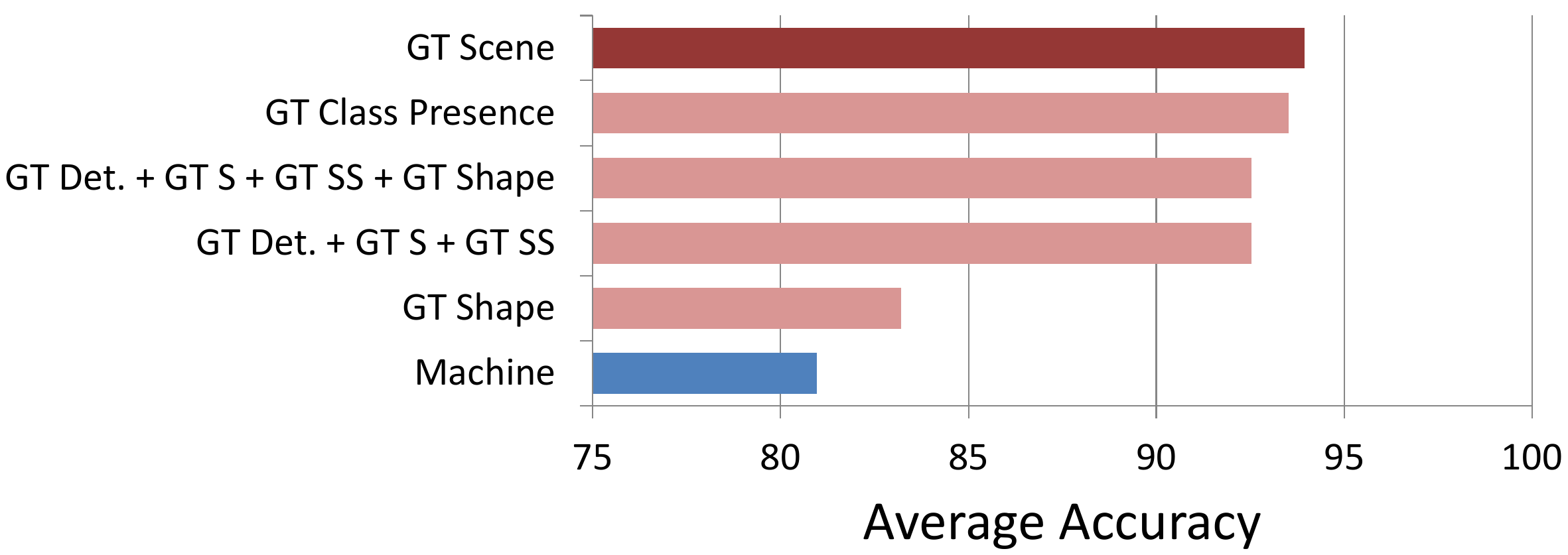}
\caption{Journey to perfection for scene classification.}
\vspace{-0.5cm}
\label{fig:scn_jour}
\end{figure}

The journey for scene recognition is shown in Figure~\ref{fig:scn_jour}. The machine performance for scene recognition is 81.0\%. Using ground truth shape potential improves the performance by 2.2\%. Using ground truth detection (Det), segment (S) and super-segment (SS) potentials instead provides 9.3\% boost in accuracy. Adding ground truth shape to this combination does not change the accuracy. Using ground truth class-presence single-handedly takes the performance to 93.5\% (an improvement of 12.5\%, while ground truth scene information is at 94.0\%). This is because the scene categories in this dataset as defined in Table~\ref{tab:msrcinfo} are object-category centric, and hence knowing whether a certain object-category is present in an image or not provides strong cues about the scene category.


\section{Conclusion} Researchers have developed sophisticated machinery for holistic scene understanding. Insights into which aspects of these models are crucial, especially for further improving state-of-the-art performance is valuable. We gather these insights by analyzing a state-of-the-art CRF model for holistic scene understanding on the MSRC dataset. 

Our analysis hinges on the use of human subjects to produce the different potentials in the model. Comparing performance of various human-machine hybrid models allows us to identify the components of the model that still have ``head room'' for improving performance. One of our findings was that human responses to local segments in isolation, while being less accurate than machines', provide complementary information that the CRF model can effectively exploit. We showed that a similar pattern of mistakes happens for the more difficult PASCAL dataset. We explored various avenues to precisely characterize this complementary nature, which resulted in a novel machine potential that significantly improves accuracy over the state-of-art. We also investigated different shape priors for the model, and it turned out that human subjects can not decipher the object shape from super-pixel boundaries any better than machines. In addition, we showed that humans can effectively leverage the contextual information incorporated in the machine model. 

We expect even more insightful findings if this model is studied on larger and more challenging datasets like the SUN dataset~\cite{s:xiao10}, which is part of future work.

{\small
\bibliographystyle{IEEEtran}      
\bibliography{all_references}
}
\vfill
\begin{biography}[{\includegraphics[width=1in,height=1in,clip,keepaspectratio]{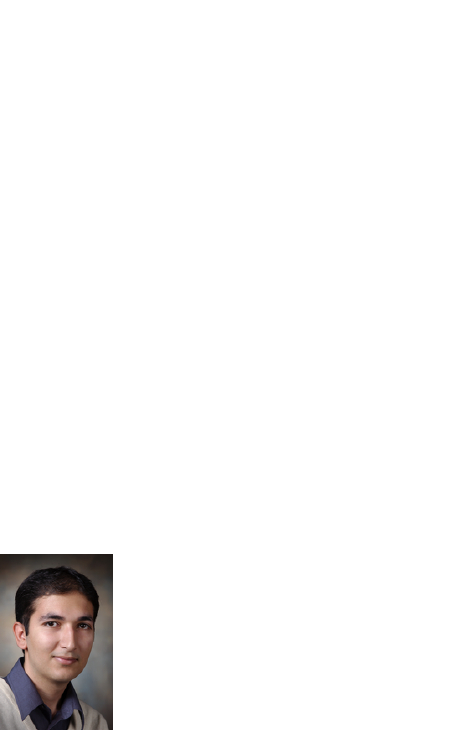}}]{Roozbeh Mottaghi} is currently a post-doctoral researcher at the Computer Science department of Stanford University. He obtained his PhD in Computer Science from UCLA in 2013. His research interests are scene understanding, object detection and segmentation, and 3D pose estimation.
\end{biography}
\begin{biography}[{\includegraphics[width=1in,height=1in,clip,keepaspectratio]{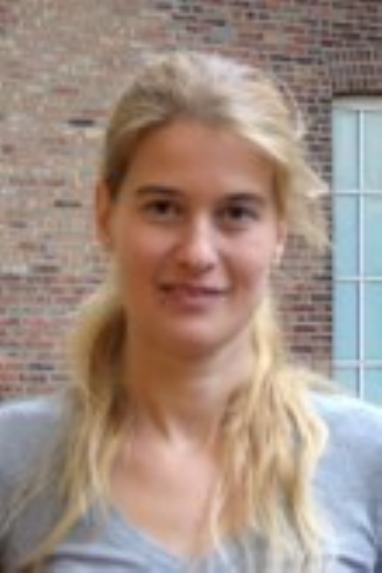}}]{Sanja Fidler} is a Research Assistant Professor at TTI-Chicago. She completed her PhD in computer science at University of Ljubljana, and was a postdoctoral fellow at University of Toronto. Her research interests are object detection and segmentation, 3D scene understanding, and combining language and vision.
\end{biography}
\begin{biography}[{\includegraphics[width=1in,height=1in,clip,keepaspectratio]{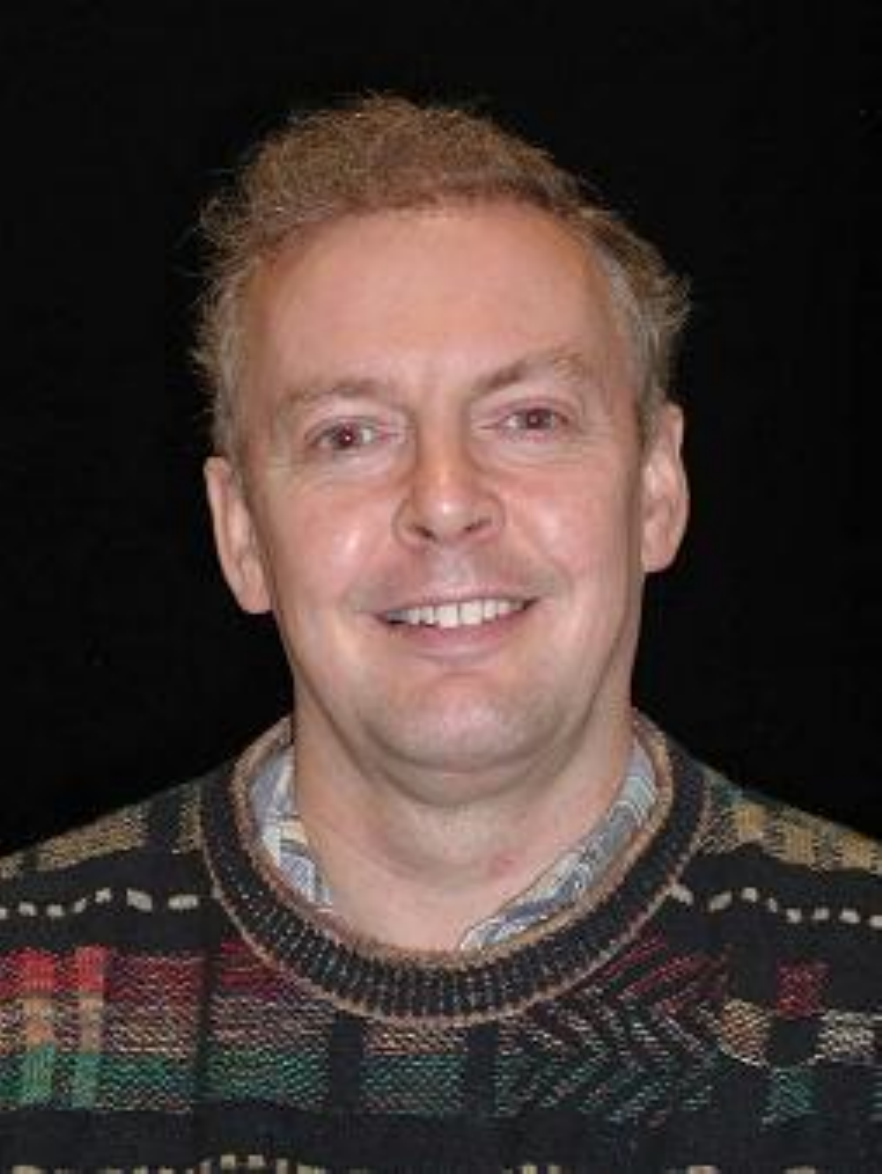}}]{Alan Yuille} received the BA degree in mathematics from the University of Cambridge in 1976. His PhD on theoretical physics, supervised by Prof. S.W. Hawking, was approved in 1981. He was a research scientist in the Artificial Intelligence Laboratory at MIT and the Division of Applied Sciences at Harvard University from 1982 to 1988. He served as an assistant and associate professor at Harvard until 1996. He was a senior research scientist at the Smith-Kettlewell Eye Research Institute from 1996 to 2002. He joined the University of California, Los Angeles, as a full professor with a joint appointment in statistics and psychology in 2002. He obtained a joint appointment in computer science in 2007. His research interests include computational models of vision, mathematical models of cognition, and artificial intelligence and neural networks.
\end{biography}
\begin{biography}[{\includegraphics[width=1in,height=1in,clip,keepaspectratio]{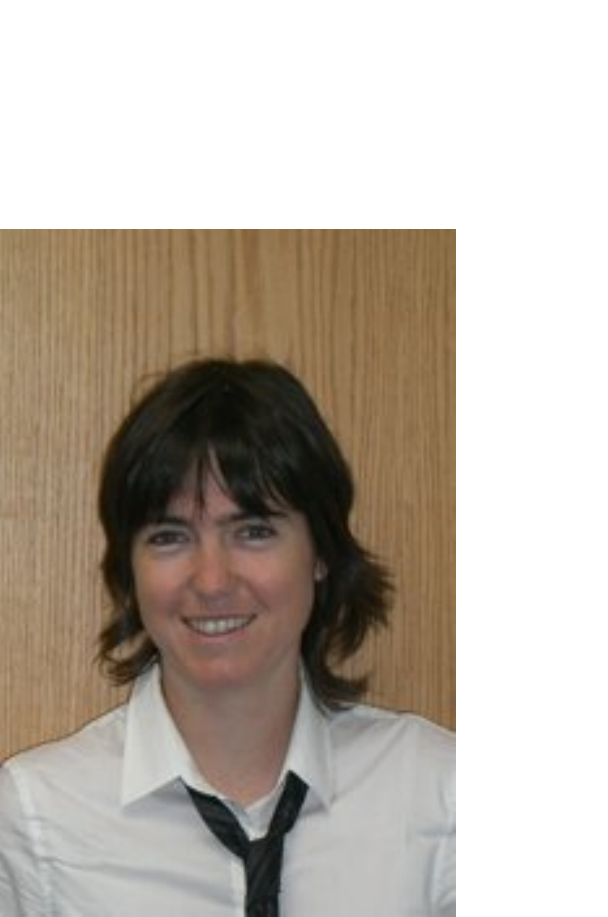}}]{Raquel Urtasun} is an Assistant
Professor at TTI-Chicago. Previously, she was a postdoctoral research scientist at UC Berkeley and ICSI and a postdoctoral associate at the Computer Science and Artificial Intelligence Laboratory (CSAIL) at MIT. Raquel Urtasun
completed her PhD at the Computer Vision Laboratory at EPFL, Switzerland. Her major interests are statistical machine learning and computer vision.
\end{biography}
\begin{biography}[{\includegraphics[width=1in,height=1in,clip,keepaspectratio]{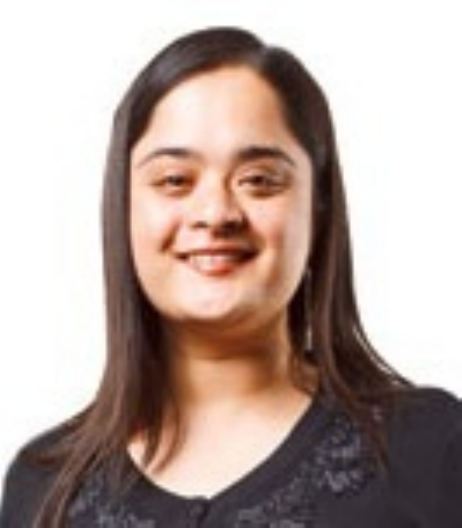}}]{Devi Parikh} is an Assistant Professor in the Bradley Department of Electrical and Computer Engineering at Virginia Tech (VT), where she leads the Computer Vision Lab. She is also a member of the Virginia Center for Autonomous Systems (VaCAS) and the VT Discovery Analytics Center (DAC). 

Prior to this, she was a Research Assistant Professor at Toyota Technological Institute at Chicago (TTIC), an academic computer science institute affiliated with University of Chicago. She has held visiting positions at Cornell University, University of Texas at Austin, Microsoft Research, MIT and Carnegie Mellon University. She received her M.S. and Ph.D. degrees from the Electrical and Computer Engineering department at Carnegie Mellon University in 2007 and 2009 respectively. She received her B.S. in Electrical and Computer Engineering from Rowan University in 2005. 

Her research interests include computer vision, pattern recognition and AI in general and visual recognition problems in particular. Her recent work involves leveraging human-machine collaborations for building smarter machines. She has also worked on other topics such as ensemble of classifiers, data fusion, inference in probabilistic models, 3D reassembly, barcode segmentation, computational photography, interactive computer vision, contextual reasoning and hierarchical representations of images. 

She was a recipient of the Carnegie Mellon Dean's Fellowship, National Science Foundation Graduate Research Fellowship, Outstanding Reviewer Award at CVPR 2012, Google Faculty Research Award in 2012 and the 2011 Marr Best Paper Prize awarded at the International Conference on Computer Vision (ICCV).
\end{biography}

\vfill

\end{document}